\documentclass[10pt,twocolumn,twoside] {IEEEtran}

\usepackage{amsmath,graphicx}
\usepackage[english]{babel}
\usepackage[latin1]{inputenc}
\usepackage{tikz}

\newcommand{\displaycomments}

\ifdefined\displaycomments
\newcommand{\note}[1]{\textcolor{red}{\emph{#1}}}
\else
\newcommand{\note}[1]{}
\fi

\usepackage{xspace}
\usepackage{amsmath}
\usepackage{amsthm}
\usepackage{amssymb}
\usepackage{graphicx}      
\usepackage{algorithm}
\usepackage{algorithmicx} 
\usepackage{cite} 
\usetikzlibrary{arrows}
\usepackage{pgfplots}

\newtheorem{mydef}{Definition}

\newtheorem{theo}{Theorem}

\theoremstyle{definition}
\newtheorem{exmp}{Example}

\newcommand{\E}{\mathop{\mathbb E}}
\newcommand{\Tr}{\operatorname{Tr}}

\renewcommand{\t}[1]{\mathrm{T}#1}
\newcommand{\N}{\mathcal{N}}
\newcommand{\ie}{i.e.,\xspace}

\newcommand{\diag}{\operatorname{Diag}}
\renewcommand{\baselinestretch}{0.9} 
\usepackage{cite} 
\usepackage{stfloats} 
\usepackage{amsmath}

\def\x{{\mathbf x}}
\def\y{{\mathbf y}}
\def\z{{\mathbf z}}

\def\x{{\mathbf x}}

\renewcommand{\d}[1]{\;\mathrm{d}#1}
\renewcommand{\t}[1]{\mathrm{T}#1}

\newcommand{\tr}{\operatorname{tr}}

\newcommand{\Poisson}{\operatorname{Poisson}}

\newtheorem{lemma}{Lemma}
\newtheorem{remark}{Remark}

\newcommand{\IW}{\mathcal{IW}}

\newcommand{\W}{\mathcal{W}}

\newcommand{\pluseq}{\stackrel{+}{=}}

\renewcommand{\d}[1]{\;\mathrm{d}#1}
\renewcommand{\t}[1]{\mathrm{T}#1}

\newcommand{\plusapprox}{\stackrel{+}{\approx}}
\newcommand{\timesapprox}{\stackrel{\times}{\approx}}

\newcommand{\FFK}{\operatorname{FFK}}

\newcommand{\VB}{\operatorname{VB}}

\newcommand{\LLL}{\operatorname{LLL}}
\newcommand{\ULL}{\operatorname{ULL}}
\usepackage{booktabs}

\newcommand{\GaussianGamma}{\operatorname{GaussianGamma}}

\newcommand{\Exp}{\operatorname{Exp}}
\newcommand{\Weibull}{\operatorname{Weibull}}

\newcommand{\Gam}{\operatorname{Gamma}}

\newcommand{\IGam}{\operatorname{IGamma}}

\usetikzlibrary{bayesnet}
\usepackage{stfloats}
\usepackage{subcaption}
\usepackage{multirow}
\usepackage{booktabs}
\usepackage{mwe}
\usepackage{graphicx}

\title{Bayesian Inference via Approximation of Log-likelihood for Priors in Exponential Family}

\author{ Tohid Ardeshiri, Umut Orguner, and Fredrik Gustafsson
        
\thanks{Tohid Ardeshiri and Fredrik Gustafsson are with the Department of Electrical Engineering, Link\"{o}ping University, 58183 Link\"{o}ping, Sweden, (e-mail: tohid@isy.liu.se, fredrik@isy.liu.se). The authors gratefully acknowledge funding from the Swedish Research Council (VR) for the project Scalable Kalman Filters.}
\thanks{ Umut Orguner is  with Department of Electrical and Electronics Engineering, Middle East Technical University, 06531 Ankara Turkey, (email: umut@metu.edu.tr).}

}

\begin{document}
\maketitle
\begin{abstract}
In this paper, a Bayesian inference technique based on Taylor series approximation of the logarithm of the likelihood function is presented. The proposed approximation is devised for the case, where the prior distribution belongs to the exponential family of distributions. The logarithm of the likelihood function is linearized with respect to the sufficient statistic of  the prior distribution in  exponential family  such that the posterior obtains the same exponential family form as the prior. Similarities between the proposed method and the extended Kalman filter for nonlinear filtering are illustrated. Furthermore, an extended target measurement update for target models where the target extent is represented by a random matrix having an inverse Wishart distribution is derived.
The approximate update covers the important case where the spread of measurement is due to the target extent as well as the measurement noise in the sensor.
\end{abstract}
\begin{keywords}
Approximate Bayesian inference, exponential family, Bayesian graphical models, extended Kalman filter, extended target tracking, group target tracking, random matrices, inverse Wishart. 
\end{keywords}
\section{Introduction}
Determination of the posterior distribution of a latent variable $\x$ given the measurements (observed data) $\y$ is at the core of Bayesian inference using probabilistic models. 
These probabilistic models  describe the relation between the random latent variables, the deterministic parameters, and the measurements. Such relations are specified  by prior distributions of the latent variables $p(\x)$, and the likelihood function $p(\y|\x)$ which give a probabilistic description of the  measurements given (some of) the latent variables.
Using the probabilistic model and measurements the exact posterior can be expressed in a functional form using the Bayes rule
\begin{align}
p(\x|\y)=\frac{p(\x)p(\y|\x)}{\int{p(\x)p(\y|\x) \d \x}}. \label{eq:Bayesrule}
\end{align}

The exact posterior distribution can be analytical. A subclass of cases where the posterior is analytical is when the posterior belongs to the same family of distributions as the prior distribution. In such cases, the prior distribution  is called a conjugate prior for the likelihood function.
A well-known example where an analytical posterior is obtained using conjugate priors is when the latent variable is \textit{a priori} normal-distributed and the likelihood function given the latent variable as its mean is again normal.
 
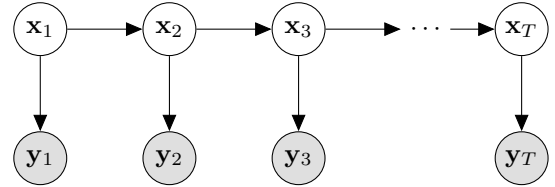
\begin{figure}[t]
  \begin{center}
    \begin{tikzpicture}
  \node[obs]                              (y1) {$\y_1$};
  \node[latent, above=of y1] 							(x1) {$\x_1$};
  \node[obs,right=of y1]                  (y2) {$\y_2$};
  \node[latent, above=of y2] 							(x2) {$\x_2$};
  \node[obs,right=of y2]                  (y3) {$\y_3$};
  \node[latent, above=of y3] 							(x3) {$\x_3$};
  \node[right=of y3]                      (y4) {$$};
  \node[right=of x3] 							        (x4) {$\mathbf{\cdots}$};
	\node[obs,right=of y4]                  (yT) {$\y_T$};
  \node[latent, above=of yT] 							(xT) {$\x_T$};                                     
  \edge {x1} {y1,x2} ; %
  \edge {x2} {y2,x3} ; %
  \edge {x3} {y3,x4} ; %
	\edge {x4} {xT} ; %
  \edge {xT} {yT} ; %
\end{tikzpicture}
  \end{center}
  \caption{A probabilistic graphical model for stochastic dynamical system with latent state $\x_k$ and measurements $\y_k$.}
	\label{fig:graphicaldynamicalsystem}  
\end{figure}

The conjugacy is especially useful when measurements are  processed sequentially as they appear in the filtering task for stochastic dynamical systems known as hidden Markov models (HMMs) whose probabilistic graphical model is presented in Fig.~\ref{fig:graphicaldynamicalsystem}. In such filtering problems, the posterior to the last processed measurement is in the same form as the prior distribution before the measurement update. Thus, the same inference algorithm can be used in a recursive manner.   

The exact posterior distribution of a latent variable can not always be given a compact analytical expression since the integral in the denominator of \eqref{eq:Bayesrule} may not be available in analytical form. Consequently, the number of parameters needed to express the posterior distribution will increase every time the Bayes rule~\eqref{eq:Bayesrule} is used for inference.  Several methods for approximate inference over probabilistic models are proposed in the literature such as variational Bayes  \cite{jordan1999, JordanG1999}, expectation propagation  \cite{Minka01}, integrated nested Laplace approximation (INLA) \cite{inla2007}, generalized linear models (GLMs)~\cite{GLM1972} and, Monte-Carlo (MC) sampling methods  \cite{Hastings70,Geman1984}.  

Variational Bayes (VB) and expectation propagation (EP)  are two analytical optimization-based solutions for the approximate Bayesian inference \cite{jordan2008}. In these two approaches,  Kullback-Leibler divergence \cite{CoverT2006} between the true posterior distribution and an approximate posterior is minimized. INLA is a technique to perform approximate Bayesian inference in latent Gaussian models \cite{tutz2001} using the Laplace approximation. GLMs are an extension of  ordinary linear regression  when errors belong to the exponential family.  

Sampling methods such as particle filters and Markov Chain Monte Carlo (MCMC) methods provide a general class of numerical solutions to the approximate Bayesian inference problem. However, in this paper, our focus is on fast analytical approximations which are applicable to large-scale inference problems.  The proposed solution is built on properties of the exponential family of distributions and earlier work on extended Kalman filter~\cite{smith1962}.

In this paper, a Bayesian inference technique based on Taylor series approximation of the logarithm of the likelihood function is presented. 
The proposed approximation  is derived for the case where the prior distribution belongs to the exponential family of distributions. The rest of this paper is organized as follows;   
In Section~\ref{sec:exponentialfamily} an introduction to exponential family of distribution is provided. In Section~\ref{sec:LLL} a general algorithm for approximate inference  in the exponential family of distributions is suggested. We show how the proposed technique is related to  the extended Kalman filter for nonlinear filtering  in Section~\ref{sec:ekf}. A novel measurement update for tracking extended targets  using random matrices is given in Section~\ref{sec:ettviaLLL}. The new measurement update for tracking extended targets is evaluated in numerical simulations in Section~\ref{sec:numsim}. The concluding remarks are given in Section~\ref{sec:conclusion}.


\section{The Exponential Family}\label{sec:exponentialfamily} 
The exponential family of distributions \cite{jordan2008} include many common distributions such as Gaussian, beta, gamma and Wishart. For $\x \in \mathcal{X}$ the exponential family in its natural form can be represented by 
\begin{equation}
p(\x;\eta)=h(\x)\exp(\eta\cdot T(\x)-A(\eta)),
\label{eq:ExponentialFamilyNaturalForm}
\end{equation}
where  $\eta$ is the natural parameter, $T(\x)$ is the sufficient statistic, $A(\eta)$ is log-partition function and  $h(\x)$ is the base measure. $\eta$ and  $T(\x)$ may be vector-valued.
Here $a \cdot b$ denotes the inner product of $a$ and $b$\footnote{In the rest of this paper we may use $a^\t{}b$  in place of $a \cdot b$ when $a$ and $b$ are vector valued. When $a$ and $b$ are matrix valued we may equivalently write $\Tr(a^\t{}b)$ instead of $\mathbf{vec}(a)\cdot \mathbf{vec}(b)$ where $\Tr(\cdot)$ is the trace operator and the operator $\mathbf{vec}(\cdot)$ vectorizes its argument (lays it out in a vector).}.
 The log-partition function is defined by the integral
\begin{align}
A(\eta)\triangleq \log \int_{\mathcal{X}}{h(\x)\exp({\eta\cdot T(\x)}) \d \x}.
\end{align}
Also, $\eta \in \Omega=\{\eta \in \mathbb{R}^m |A(\eta)< +\infty\}$ where $\Omega$ is the natural parameter space. Moreover, $\Omega$ is a convex set and $A(\cdot)$ is a convex function on $\Omega$.

When a probability density function (PDF) in the exponential family is parametrized by a non-canonical parameter $\theta$, the PDF in \eqref{eq:ExponentialFamilyNaturalForm} can be written as  
\begin{equation}
p(\x;\theta)=h(\x)\exp(\eta(\theta)\cdot T(\x)-A(\eta(\theta))),
\label{eq:ExponentialFamilythetaForm}
\end{equation}
where $\theta$ is the non-canonical parameter vector of the PDF.

\begin{exmp} \label{ex:exponentialfamily} Consider the normal distribution $p(\x)=\N(\x;\mu,\Sigma)$ with mean $\mu\in\mathbb{R}^d$ and covariance matrix $\Sigma$. Its PDF can be written in exponential family form given in \eqref{eq:ExponentialFamilyNaturalForm} where
\begin{subequations}
\label{eq:normalExpForm}
\begin{align}
h(\x)&=(2\pi)^{-\frac{d}{2}},\\
T(\x)&=\begin{bmatrix} \x\\\mathbf{vec}(\x\x^\t) \end{bmatrix}\label{eq:sufstats},\\
\eta(\mu,\Sigma)&=\begin{bmatrix} \Sigma^{-1}\mu\\\mathbf{vec}(-\frac{1}{2}\Sigma^{-1}) \end{bmatrix}\label{eq:Natparams},\\
A(\eta(\mu,\Sigma))
&=\frac{1}{2}\mu^\t \Sigma^{-1}\mu+\frac{1}{2}\log|\Sigma|.
\end{align} 
\end{subequations}\hfill$\blacksquare$ 
\end{exmp}

\begin{remark}
In the rest of this document, the $\mathbf{vec}(\cdot)$ operators will be  dropped to keep the notation less cluttered. That is, we will use the simplified notation $T(\x)=(\x,\x\x^\t)$ instead of \eqref{eq:sufstats} and $\eta(\mu,\Sigma)=(\Sigma^{-1}\mu,-\frac{1}{2}\Sigma^{-1})$ instead of \eqref{eq:Natparams}. 
\end{remark}

\subsection{Conjugate Priors in Exponential Family}\label{sec:conjprior}
In this section we study the conjugate prior family for a likelihood function belonging to the exponential family.
Consider  $m$ independent and identically distributed (IID)  measurements $\mathcal{Y}\triangleq\{ \y^j\in \mathbb{R}^d | 1\leq j \leq m\}$ and the likelihood function $p(\mathcal{Y}|\x,\lambda)$ in exponential family, where $\lambda$ is a set of hyper-parameters of the likelihood and $\x$ is the latent variable. The likelihood of the $m$ IID measurements can be written as
\begin{align}
&p(\mathcal{Y}|\x,\lambda)=\Big(\prod_{j=1}^{m}h(\y^j)\Big)\nonumber\\
&\hspace{4mm}\times\exp\left(\eta(\x,\lambda) \cdot \sum_{j=1}^m T(\y^j)-m A(\eta(\x,\lambda))\right). \label{eq:ExponentialFamilyLikelihood}
\end{align}

We seek a conjugate prior distribution for $\x$ denoted by $p(\x)$. The prior distribution $p(\x)$ is a conjugate prior if the posterior distribution 
\begin{align}
p(\x|\mathcal{Y}) \propto p(\mathcal{Y}|\x,\lambda)p(\x)\label{eq:ExponentialFamilyPosterior}
\end{align}
is also in the same exponential family as the prior.
Let us assume that the prior is in the form
\begin{align}
p(\x)\propto \exp (\mathcal{F}(\x))
\end{align} 
for some function $\mathcal{F}(\cdot)$.
Hence, 
\begin{align}
&p(\x|\mathcal{Y}) \propto p(\mathcal{Y}|\x,\lambda)\exp (\mathcal{F}(\x))\\
&\propto  \exp\left(\eta(\x,\lambda) \cdot \sum_{j=1}^m T(\y^j)-m A(\eta(\x,\lambda))+\mathcal{F}(\x)\right).\label{eq:ExponentialFamilyPosterior-expanded}
\end{align}
For the posterior \eqref{eq:ExponentialFamilyPosterior-expanded} to be in the same exponential family as the prior \cite{LectureMJ}, $\mathcal{F}(\cdot)$ needs to be in the form
\begin{align}
\mathcal{F}(\x)= \rho_1\cdot\eta(\x,\lambda) - \rho_0 A(\eta(\x,\lambda))
\end{align} 
for some $\rho\triangleq(\rho_0, \rho_1)$, such that 
\begin{align}
&p(\x|\mathcal{Y}) \propto\nonumber\\
& \exp\left(\left(\rho_1+ \sum_{j=1}^m T(\y^j)\right)\cdot\eta(\x,\lambda) -(m+\rho_0) A(\eta(\x,\lambda))\right).\label{eq:ExponentialFamilyPosteriorLinearcombination}
\end{align}
Hence, the conjugate prior for the likelihood \eqref{eq:ExponentialFamilyLikelihood} is parametrized by $\rho$ and is given by 
\begin{align}
p(\x;\rho)=\frac{1}{Z}\exp\left(   \rho_1 \cdot \eta(\x,\lambda) - \rho_0 A(\eta(\x,\lambda) \right),
\end{align} 
where
\begin{align}
Z={\int{\exp\left(   \rho_1 \cdot \eta(\x,\lambda) - \rho_0 A(\eta(\x,\lambda) \right) \d \x}}.
\end{align} 

In Example \ref{ex:conjprior}, the conjugate prior for the normal likelihood is derived using the exponential family form for the normal distribution given in Example~\ref{ex:exponentialfamily}. In Example \ref{ex:conjsin}, a conjugate prior for a more complicated likelihood function is derived.

\begin{exmp}\label{ex:conjprior} Consider the measurement $\y$ with the likelihood  $p(\y|\x)=\N(\y;C \x,R)$. The likelihood can be written in exponential family form as in
\begin{align}
p(\y|\x)=(2\pi)^{-\frac{d}{2}}\exp{\left(\eta(C\x,R)\cdot T(\y) -A(\eta(C\x,R))\right)}
\end{align} 
where $T(\y)=(\y,\y \y^\t)$, $\eta(C\x,R)=(R^{-1}C \x,-\frac{1}{2}R^{-1})$ and 
\begin{subequations}
\begin{align}
A(\eta(C\x,R))&=\frac{1}{2}\x^\t C^\t R^{-1}C \x+\frac{1}{2}\log|R|\\
&=\frac{1}{2}\left( C^\t R^{-1}C\right) \cdot \x \x^\t+\frac{1}{2}\log|R|.
\end{align}
\end{subequations} 
Therefore, the likelihood function can be written in the form
\begin{align}
p(\y|\x)\propto 
&\exp\left(  \left( C^\t R^{-1} \y, - \frac{1}{2}C^\t R^{-1}C\right) \cdot T(\x)\right).
\end{align} 
Hence, the conjugate prior for the likelihood function of this example should have $T(\cdot)$ as its sufficient statistic  \ie the conjugate prior is normal. When the prior distribution is normal \ie $p(\x)=\N(\x;\mu,\Sigma)$ the posterior distribution obeys
\begin{align}
&p(\x|\y)\propto\nonumber\\
&\exp\left(  \left( C^\t R^{-1} \y+ \Sigma^{-1}\mu ,- \frac{1}{2}C^\t R^{-1}C -\frac{1}{2}\Sigma^{-1}\right) \cdot T(\x)\right)
\end{align}
which is the information filter form for the Kalman filter's measurement update \cite{Kalman60}.\hfill$\blacksquare$ 
\end{exmp}

\begin{exmp}\label{ex:conjsin} Consider the measurement $\y$ with the likelihood  
\begin{align}
p(\y|\x)\propto\exp(-\frac{1}{24}(\y-\x)^2+\cos(\y-\x)).\label{eq:likeconjsin}
\end{align} 
The likelihood function is a multi-modal likelihood function and is illustrated in Fig.~\ref{fig:conjsin} for $\y=3$. 
The likelihood can be written in the form
\begin{align}
p(\y|\x)\propto\exp( \lambda \cdot T(\x))\label{eq:explikelihoodconjsin}
\end{align} 
where $\lambda=\begin{bmatrix} -\frac{1}{24} & \frac{\y}{12} & \cos \y &\sin{\y} \end{bmatrix}^\t$ and $T(\x)=\begin{bmatrix} \x^2& \x & \cos \x & \sin \x \end{bmatrix}^\t$. 
Hence, a conjugate prior for the likelihood function of this example should have $T(\cdot)$ as its sufficient statistic such as $p(\x)\propto \exp\left(\begin{bmatrix} -\frac{1}{10} &0& 0& 0 \end{bmatrix}T(\x)\right)$. Thus, the posterior for the likelihood function and the prior distribution obeys
\begin{align}
&p(\x|\y)\propto\exp\left(  \begin{bmatrix} -\frac{1}{24}-\frac{1}{10} & \frac{\y}{12} & \cos \y &\sin{\y} \end{bmatrix}^\t \cdot T(\x)\right).\label{eq:postconjsin}
\end{align}
Although such posteriors can be computed analytically up to a normalization factor, the density may not be available in analytical form. The numerically normalized prior and the posterior are illustrated in Fig.~\ref{fig:conjsin}.	
\begin{figure}[ht]
\centering
\includegraphics[width=.45\textwidth]{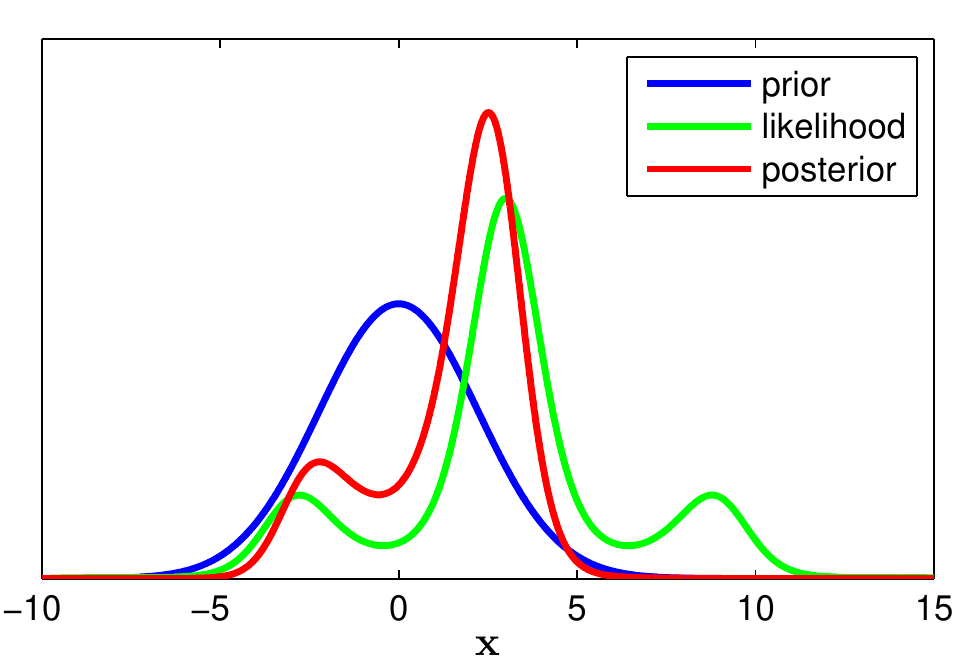}			
\caption{The likelihood function \eqref{eq:likeconjsin} prior distribution \eqref{eq:explikelihoodconjsin} and the posterior distribution \eqref{eq:postconjsin} in Example~\ref{ex:conjsin} are plotted.}
\label{fig:conjsin}	
\end{figure}
\hfill$\blacksquare$
\end{exmp}

Similar reasoning as the one presented in this section can be applied to the case where  the prior is fixed and the likelihood function is to be selected; The concept of  ``conjugate likelihoods" will be considered in the sequel.
\subsection{Conjugate Likelihoods in Exponential Family}
First, we define the conjugate likelihood functions;
\begin{mydef}
A likelihood  function $p(\y|\x)$ is called conjugate likelihood for prior distribution $p(\x)$ in a family of distributions, if the posterior distribution $p(\x|\y)$ belongs to the same family of distributions as the prior distribution.
\end{mydef}
Now, consider the prior distribution $p(\x;\rho)$ in the exponential family form on the latent variable $\x\in\mathcal{X}$, where $\rho$ is a set of  hyper-parameters for the prior
\begin{align}
p(\x;\rho)=h(\x)\exp\left(\eta(\rho) \cdot T(\x)- A(\eta(\rho))\right). \label{eq:ExponentialFamilyPrior}
\end{align}
 We will follow the treatment given for characterization of the conjugate priors in Section~\ref{sec:conjprior} as well as~\cite{LectureMJ}, for the conjugate likelihood functions.
Assume now that we have observed  $m$ IID measurements $\mathcal{Y}\triangleq\{ \y^j\in \mathbb{R}^d | 1\leq j \leq m\}$.
Let the likelihood of the measurements  be written as
\begin{align}
p(\mathcal{Y}|\x)\propto \exp\left(\mathcal{L}(\x,\mathcal{Y})\right)\label{eq:generalLikelihood}
\end{align}
for some function $\mathcal{L}(\cdot)$.
We seek a conjugate likelihood function in the exponential family. Hence, the posterior distribution 
has to be in the exponential family where,
\begin{align}
p(\x|\mathcal{Y}) \propto\ & h(\x)\exp\left(\eta(\rho) \cdot T(\x)- A(\eta(\rho))+\mathcal{L}(\x,\mathcal{Y})\right).\label{eq:ExponentialFamilyPosterior-conjLikeexpanded}
\end{align}
For the posterior \eqref{eq:ExponentialFamilyPosterior-conjLikeexpanded} to be in the same exponential family, $\mathcal{L}(\cdot)$ needs to be in the form
\begin{align}
\mathcal{L}(\x,\y)\pluseq \lambda(\mathcal{Y})\cdot T(\x)
\end{align} 
where $\pluseq$ means equality up to an additive constant with respect to latent variable $\x$  such that 
\begin{align}
p(\x|\mathcal{Y}) \propto\ & h(\x)\exp\left((\eta(\rho)+\lambda(\mathcal{Y})) \cdot T(\x)\right)
.\label{eq:ExponentialFamilyPosteriorLinearcombinationConjLikelihood}
\end{align}
Hence, the conjugate likelihood for the prior distribution family \eqref{eq:ExponentialFamilyPrior} is parametrized by $\lambda(\mathcal{Y})$ and is given by
\begin{align}
p(\mathcal{Y}|\x)\propto\exp\left(  \lambda(\mathcal{Y}) \cdot T(\x) \right).\label{eq:necessarycondition}
\end{align}
\begin{table}[t]
\caption{Some continuous exponential family distributions and their sufficient statistic are listed. }
\centering
\begin{tabular}{l|c}
\toprule
Continuous Exp. Family Distribution  & $T(\cdot)$ \\
\hline \rule{-2pt}{2ex} 
Exponential distribution & $\x$\\
Normal distribution with known variance $\sigma^2$& $\x/\sigma$\\ 
Normal distribution & $(\x,\x\x^\t)$\\
Pareto distribution with known minimum $x_m$ &$\log \x$\\
Weibull distribution with known shape $k$ & $\x^k$\\
Chi-squared distribution & $\log \x$\\
Dirichlet distribution& $(\log \x_1,\cdots,\log\x_n) $\\
Laplace distribution with known mean $\mu$ &  $|\x-\mu|$\\
Inverse Gaussian distribution& $(\x,1/\x)$\\
Scaled inverse Chi-squared distribution & $(\log\x, 1/\x)$ \\
Beta distribution& $(\log\x,\log(1-\x))$\\
Lognormal distribution & $(\log \x,(\log\x)^2)$\\
Gamma distribution & $(\log \x, \x)$ \\
Inverse gamma  distribution& $(\log \x,1 / \x)$   \\
Gaussian Gamma distribution& $(\log \tau,\tau,\tau \x ,\tau \x^2)$ \\
Wishart  distribution & $(\log|X|, X) $ \\
Inverse Wishart distribution & $(\log|X|, X^{-1}) $ \\
\bottomrule
\end{tabular}
\label{table:sufficientstatistic}
\end{table}
The property expressed in \eqref{eq:necessarycondition} is the necessary condition which should hold for a conjugate likelihood for a given prior distribution family. Another condition which should hold is that the log-partition function of the posterior \eqref{eq:ExponentialFamilyPosteriorLinearcombinationConjLikelihood} should obey
\begin{align}
A= \log \int_{\mathcal{X}}{h(\x)\exp({(\eta(\rho)+\lambda(\mathcal{Y}) )\cdot T(\x)}) \d \x} < \infty
\end{align}
such that the natural parameter of the posterior belongs to the natural parameter space $\Omega$.
These properties of the conjugate likelihood are summarized in theorem~\ref{theorem:iffconjugatelikelihood}.
\begin{theo}\label{theorem:iffconjugatelikelihood}
The likelihood  function $p(\y|\x)$ is a conjugate likelihood for the prior distribution in an exponential family $p(\x)=h(\x)\exp(\eta\cdot T(\x)-A(\eta)) $ with $\x \in\mathcal{X}$  if and only if
\begin{enumerate}
\item $ p(\y|\x)\propto \exp(\lambda\cdot T(\x))$ for some $\lambda$ and,
\item  the likelihood function $p(\y|\x)$ is integrable with respect to $\y$ and,
\item  $\log \int_{\mathcal{X}}{h(\x)\exp({(\eta+\lambda)\cdot T(\x)}) \d \x} < \infty$ for $\forall \eta \in \Omega$.
\end{enumerate} 
\hfill$\blacksquare$
\end{theo}

In Examples \ref{ex:conjlikelihood} and~\ref{ex:conjsin-likelihood}, the conjugate likelihood functions for two prior distributions are derived using the exponential family form. 

\begin{exmp} \label{ex:conjlikelihood} Consider the normal distribution $p(\x)=\N(\x;\mu,\Sigma)$ whose PDF is written in exponential family form in Example~\ref{ex:exponentialfamily}. The conjugate likelihood  function has to be in the form 
\begin{align}
p(\y|\x)\propto\exp\left(\lambda_1(\y) \cdot \x + \lambda_2(\y) \cdot \mathbf{vec}(\x \x^\t) \right)
\end{align} 
for some permissible $\lambda(\cdot)\triangleq [\lambda_1,\lambda_2]$,
which includes $p(\y|\x)=\N(\y;C\x,R)$ for a matrix $C$ with appropriate dimensions and symmetric $R \succ 0$.\hfill$\blacksquare$
\end{exmp}
\begin{exmp}\label{ex:conjsin-likelihood} 
Consider  the prior distribution 
\begin{align}
&p(\x)\propto\exp\left(  \lambda \cdot T(\x)\right)\label{eq:priorconjsin}
\end{align}
where $T(\x)=\begin{bmatrix} \x^2& \x & \cos \x & \sin \x \end{bmatrix}^\t$. 
The family of conjugate likelihood density functions $p(\y|\x)$ is parametrized by $\psi \in \mathbb{R}^4$ such that 
\begin{enumerate}
\item  $\log p(\y|\x)\pluseq \psi\cdot T(\x)$,
\item  $\int_{-\infty}^{\infty}{p(\y|\x)\d \y }=1$, \ie the likelihood is integrable with respect to $\y$,
\item  $\int_{-\infty}^{\infty}{\exp({(\lambda+\psi)\cdot T(\x)}) \d \x} < \infty$, \ie the posterior is integrable with respect to $\x$.
\end{enumerate}
A member of this family is 
\begin{align}
p(\y|\x)\propto\exp(-\alpha^2(\y-\x)^2+\cos(\beta\y-\x+\gamma))
\end{align} 
for $[\alpha,\beta,\gamma]^\t\in\mathbb{R}^3$. \hfill$\blacksquare$
\end{exmp} 


In Table~\ref{table:sufficientstatistic} the sufficient statistic for some continuous members of the exponential family are given. Some members of the continuous exponential family have  conjugate likelihoods which are summarized in Table~\ref{table:conjugatelikelihood}.

\begin{table*}[ht]
\caption{Some prior distributions and their conjugate likelihood functions are listed. The arguments of the PDFs on the left columns are the latent random variables. On the right hand side column the conjugate likelihood functions $p(\y|\cdot)$ are given where $\y$ denotes the measurement. In the middle column the sufficient statistic functions of the prior distributions in the left column are given. The logarithm of the conjugate likelihood functions are linear in sufficient statistic function to their left.}
\centering
\begin{tabular}{l|c|l}
\toprule
Prior distribution with sufficient statistic $T(\cdot)$ & $T(\cdot)$ & Conjugate likelihood function\\
\hline \rule{-2pt}{3ex} 
$\N(\x;\mu;\Sigma)$    & $(\x,\x\x^\t)$   & $\N(\y;C\x,R)\propto \exp\left(\Tr \left(R^{-1}(\y-C\x)(\y-C\x)^\t\right)\right) $\\
$\N(\x;\mu;\Sigma)$    & $(\x,\x\x^\t)$   & $\log-\N(\y;\x,\sigma^2)\propto \exp\left(-\frac{1}{2\sigma^2}(\log(\y)-\x)^2\right) $\\
$\Gam(\x;\alpha,\beta)$& $(\log \x, \x)$  & $\Exp(\y;\x)=\x\exp(-\x\y)$  \\
$\Gam(\x;\alpha,\beta)$& $(\log \x, \x)$  & $\IGam(\y;\alpha^\prime,\x)\propto \x^{\alpha^\prime} \exp(-\x/\y)$  \\
$\Gam(\x;\alpha,\beta)$& $(\log \x, \x)$  & $\Gam(\y;\alpha^\prime,\x)\propto \x^{\alpha^\prime} \exp(-\x\y)$  \\
$\Gam(\x;\alpha,\beta)$& $(\log \x, \x)$  & $\N(\y;\mu,\x^{-1})\propto \x^{\frac{1}{2}}\exp(-\frac{\x}{2}(\y-\mu)^2) $\\
$\IGam(\x;\alpha;\beta)$    & $(\log \x,1 / \x)$   & $\N(\y;\mu,\x)\propto \x^{-\frac{1}{2}}\exp(-\frac{1}{2\x}(\y-\mu)^2) $\\
$\IGam(\x;\alpha;\beta)$    & $(\log \x,1 / \x)$   & $\Weibull(\y;\x,k)\propto \frac{k}{\x}\y^{k-1} \exp\left(-\frac{\y^k}{\x} \right) $\\
$\GaussianGamma(\x, \tau ; \mu, \lambda, \alpha, \beta)$&$(\log \tau,\tau,\tau \x ,\tau \x^2)$ & $\N(\y;\x,\tau^{-1})\propto \tau^{\frac{1}{2}}\exp(-\frac{1}{2}\tau(\y-\x)^2)$\\
$\W(X;n,V)$        & $(\log|X|, X) $ & $\N(\y;\mu,X^{-1})\propto |X|^{\frac{d}{2}} \exp(\Tr( X(\y-\mu)(\y-\mu)^\t))$\\
$\IW(X;\nu,\Psi)$  & $(\log|X|, X^{-1}) $ & $\N(\y;\mu,X)\propto |X|^{-\frac{d}{2}}  \exp(\Tr( X^{-1}(\y-\mu)(\y-\mu)^\t))$\\
\bottomrule
\end{tabular}
\label{table:conjugatelikelihood}
\end{table*}

\section {Measurement Update via Approximation of the Log-likelihood } \label{sec:LLL}
In this section a method is proposed for the analytical approximation of complex likelihood functions based on linearization of the log-likelihood with respect to sufficient statistic function $T(\cdot)$. We will use the property of conjugate likelihood functions \ie linearity of log-likelihood function with respect to sufficient statistic of the prior distribution  to come up with an approximation of the likelihood function for which there exists analytical posterior distribution.

In the proposed method first, the log-likelihood function is derived in analytical form. Second, the likelihood function is approximated by a dot product of a statistic which depends on the measurement and the sufficient statistic of the prior distribution.  
For example, consider a likelihood function 
\begin{align}
p(\mathcal{Y}|\x)\propto \exp\left(\mathcal{L}(\x,\mathcal{Y})\right)
\end{align}
where the log-likelihood function is given by  $\mathcal{L}(\cdot)$ and the prior distribution is given by
\begin{align}
p(\x)= h(\x)\exp\left(\eta\cdot T(\x)-A(\eta)\right)
\end{align}
where $\eta\in\Omega$.
 If we approximate $\mathcal{L}(\cdot)$ such that
\begin{align}
\mathcal{L}(\x,\mathcal{Y})\approx \widehat{\mathcal{L}}(\x,\mathcal{Y})\pluseq  \lambda(\mathcal{Y}) \cdot T(\x)
\end{align}
for some $\lambda(\cdot)$ such that for any measurement $\mathcal{Y}$ belonging to the support of the likelihood function 
\begin{align}
\eta+\lambda(\mathcal{Y})\in \Omega
\end{align}
and, 
\begin{align}
\int \exp \left(\widehat{\mathcal{L}}(\x,\mathcal{Y})\right) \d \mathcal{Y}< \infty,
\end{align}
 then the approximate likelihood will be  conjugate  likelihood for the prior distribution with sufficient statistic $T(\x)$.

When the prior distribution is normal, the proposed approximation can be obtained using well-known Taylor series approximation at the global maximum of the log-likelihood function. This is due to the fact that $T(\x)=(\x,\x\x^\t)$ for the normal prior and a second order Taylor series approximation of the log-likelihood approximates the log-likelihood with a function linear in $T(\x)$. This mathematical convenience is used in the well-known approximate Bayesian inference technique INLA \cite{inla2007}. 
Similarly, when the prior distribution is exponential distribution, a first order Taylor series approximation can be used to approximate the log-likelihood with a  function linear in $T(\x)=\x$. 

Taylor series expansion of functions will be used in the proposed approximations in the following. Hence, we establish the notation used in this paper for Taylor series expansion of arbitrary functions before we proceed.

\subsection{Taylor series expansion}
The Taylor series expansion of  a real-valued function $f(\x)$ that is infinitely differentiable at a real vector  $\widehat{\x}$  is given by the series
\begin{align}
f(\x)=&f(\widehat{\x})+\frac{1}{1!}\nabla_\x f(\widehat{\x})\cdot (\x-\widehat{\x})\nonumber\\
&+\frac{1}{2!}\nabla_\x^2 f(\widehat{\x})\cdot \left((\x-\widehat{\x})(\x-\widehat{\x})^\t\right)+\cdots
\end{align}
where $\nabla_\x f(\widehat{\x})$ is the gradient vector which is composed of the partial
derivatives of $f(\x)$ with respect to  elements of $\x$, evaluated at $\widehat{\x}$. Similarly, $\nabla_\x^2f(\widehat{\x})$ is the Hessian matrix of $f(\x)$ containing second order partial derivatives and evaluated at $\widehat{\x}$.

When $f(\x)$ is a vector-valued function, \ie $f(\x)\in\mathbb{R}^d$ with $i$th element denoted by $f_i(\x)$, first the Taylor series expansion is derived for each element separately. Then, the expansion of $f(\x)$ at $\widehat{\x}$ is constructed by rearranging the terms in a matrix.
\begin{align}
f(\x)=&f(\widehat{\x})+\frac{1}{1!}\left(\nabla_\x f(\widehat{\x})\right)^\t (\x-\widehat{\x})\nonumber\\
&+\frac{1}{2!}\sum_{i=1}^d e_i \nabla_\x^2 f_i(\widehat{\x})\cdot \left((\x-\widehat{\x})(\x-\widehat{\x})^\t\right)+\cdots\  \label{eq:Taylorvector}
\end{align}   
In \eqref{eq:Taylorvector}, $\nabla_\x f(\widehat{\x})$ is the Jacobian matrix of $f(\x)$   evaluated at $\widehat{\x}$ and $e_i$ is a unity vector with its $i$th element equal to $1$.

\subsection{The extended Kalman filter}
\label{sec:ekf}
A well-known problem where a special case of the proposed inference technique is used is the filtering problem for nonlinear state-space models in presence of additive Gaussian noise which will be described here.
Consider the discrete-time stochastic nonlinear dynamical system 
\begin{subequations}
\begin{align}
\y_k|\x_k & \thicksim \N(\y_k;c(\x_k),R_k),\\
\x_{k+1}|\x_k& \thicksim \N(\x_{k+1};f(\x_k),Q_k),
\end{align}
\end{subequations}
where $\x_k$  and $\y_k$ are the latent variable and  the measurement at time index $k$ and $c(\cdot)$ and $f(\cdot)$ are nonlinear functions. 
A common solution to the inference problem is the extended Kalman filter (EKF) \cite{smith1962}.

In the following example, we show that the measurement update for EKF is a special case of the proposed algorithm.  

\begin{exmp}\label{ex:LLLEKF}  Consider the latent variable $\x$ with \textit{a priori} PDF $p(\x)=\N(\x;\mu,\Sigma)$ and the measurement $\y$ with the likelihood  function $p(\y|\x)=\N(\y;c(\x),R)$. The likelihood function can be written in exponential family form as  in 
\begin{align}
p(\y|\x)=(2\pi)^{-\frac{d}{2}}\exp{\left(\eta(\x,R)\cdot T(\y) -A(\eta(\x,R))\right)}
\end{align} 
where  $T(\y)=(\y,\y \y^\t)$, $\eta(\x,R)=(R^{-1}c(\x),-\frac{1}{2}R^{-1})$ and 
\begin{align}
A(\eta(\x,R))&=\frac{1}{2}\Tr \left( R^{-1}c(\x)c(\x)^\t\right)+\frac{1}{2}\log|R|.
\end{align} 
Thus,
\begin{align}
&p(\y|\x)\propto \exp\left(  R^{-1}c(\x)\cdot \y- \frac{1}{2} c(\x)^\t R^{-1}c(\x)\right)
\end{align} 
and the log-likelihood function can be written as
\begin{align}
&\mathcal{L}(\x)\pluseq  \y^\t R^{-1}c(\x) - \frac{1}{2}  c(\x)^\t R^{-1}c(\x) \label{eq:LLEKF}
\end{align} 
where $\pluseq$ means equality up to an additive constant with respect to the latent variable $\x$. 
The second order Taylor series approximation of any function will be linear in the sufficient statistic $T(\x)=(\x,\x\x^\t)$. However, such an approximation   
does not guarantee the integrability of the approximate likelihood and the corresponding approximate posterior due to the dependence of second element of the posterior's natural parameter on the measurement $\y$. 

A common solution to the log-likelihood linearization problem has been the  first order Taylor series approximation of $c(\x)$ about $\widehat{\x}=\mu$ (instead of second order Taylor series approximation of $\mathcal{L}(\x)$), \ie
\begin{align}
c(\x)\approx c(\mu)+ c^\prime(\mu) (\x-\mu)
\end{align}
where 
\begin{align}
c^\prime(\mu)\triangleq \left. \nabla_\x c(\x)\right|_{\x=\mu}.
\end{align}
 With this approximation the approximate log-likelihood becomes
\begin{align}
\mathcal{L}(\x)\plusapprox &(c^\prime(\mu)  R^{-1} (\y^\t-c(\mu)-c^\prime(\mu)^\t\mu))\cdot\x\nonumber\\
&-    \frac{1}{2} c^\prime(\mu) R^{-1} c^\prime(\mu)^\t\cdot \x\x^\t.
\end{align} 
Hence, a conjugate likelihood function for prior distribution is obtained by approximating the log-likelihood function by a function that is linear in sufficient statistic of the prior distribution. Also note that the posterior distribution obeys 
\begin{align}
p(\x|\y)\propto&\exp\left(\phi \cdot T(\x)\right)
\end{align}
where
\begin{align}
\phi&=\biggl(c^\prime(\mu)  R^{-1} (\y^\t-c(\mu)-c^\prime(\mu)^\t\mu) +\Sigma^{-1}\mu\ , \nonumber\\
&\hspace{3cm}    -    \frac{1}{2} c^\prime(\mu) R^{-1} c^\prime(\mu)^\t-     \frac{1}{2}\Sigma^{-1} \biggr)  
\end{align}
which is the extended information filter form for the extended Kalman filter~\cite[Sec.3.5.4 ]{thrun2005}. An advantage of this solution over Taylor series approximation of the log-likelihood is that the natural parameter of the posterior will be in natural parameter space by construction \ie $\frac{1}{2} c^\prime(\mu) R^{-1} c^\prime(\mu)^\t+     \frac{1}{2}\Sigma^{-1}\succ 0$.  
\hfill$\blacksquare$
\end{exmp}
\subsection{A general linearization guideline}
The approximation of a general scalar function $L(\cdot)$ with another function $\widehat{L}(\cdot)$ linear in a general vector valued function $t(\cdot)$ \ie  $\widehat{L}(\cdot)\triangleq\lambda\cdot t(\cdot)$  can be expressed as an optimization problem where various cost functions are suggested and minimized. Study of such procedures is considered outside the scope of this paper and a subject of future research. However, a few tricks are introduced here which can be used for the purpose of linearization. 


\begin{lemma} \label{lem:invertible-t}
Let $L(\x)$ be an arbitrary scalar-valued continuous and differentiable function and $t(\x)$ be a vector-valued invertible function with independent continuous elements  such that $\x=t^{-1}(\z)$ and $L(t^{-1}(\z))$ is differentiable with respect to $\z$. Then, $L(\x)$ can be linearized with respect to $t(\x)$ about $t(\widehat{\x})$ such that
\begin{align}
L(\x)\approx L(\widehat{\x})+ \Phi \cdot (t(\x)-t(\widehat{\x}))
\end{align} 
and $\Phi=\left.\nabla_{\z}L(t^{-1}(\z))\right|_{\z=t(\widehat{\x})} $.

\begin{proof}
Let $\z=t(\x)$ and $Q(\z)\triangleq L(t^{-1}(\z))$. Then, $\x=t^{-1}(\z)$ and $Q(\z)=L(\x)$ for $\z=t(\x)$. Using the first-order Taylor series approximation of the scalar function $Q(\z)$ about $\widehat{\z}\triangleq t(\widehat{\x})$ we obtain
\begin{align}
Q(\z)\approx Q(\widehat{\z})+ \left.\nabla_{\z}Q(\z)\right|_{\z=\widehat{\z}} \cdot (\z-\widehat{\z}).
\end{align} 
Hence, the proof follows.
\end{proof}
\end{lemma}
\begin{remark}
The variable $\z$ and the function $t(\x)$ can be scalar-valued, vector-valued or even matrix-valued. Further,
$\nabla_{\z}Q(\z)$ can be computed using the chain rule and the fact that $Q(\z)=L\left(t^{-1}(\z)\right)$.
\end{remark}

%
%
%
%
%

The elements of sufficient statistic of most continuous members of the exponential family are dependent, such as $\x$ and $\x\x^\t$ for normal distribution. Furthermore, some of them are not invertible such as $\log|X|$ for Wishart distribution.  However, the linearization method given in Lemma \ref{lem:invertible-t} can be applied to individual summand terms of  the log-likelihood function and individual elements of the sufficient statistic. Due to the freedom in choosing the element of the sufficient statistic to linearize the log-likelihood with respect to, there is no unique solution for the linearization problem.  In Example~\ref{ex:normal-igamma} we will use Lemma~\ref{lem:invertible-t} and linearize a log-likelihood function in various ways and describe their properties.

\begin{exmp}\label{ex:normal-igamma}  Consider the scalar latent variable $\x$ with \textit{a priori} PDF $p(\x)=\IGam(\x;\alpha,\beta)$ and the measurement $\y$ with the likelihood  function $p(\y|\x)=\N(\y;0,\x+\sigma^2)$. Since the exact posterior density is not analytically tractable, an approximation is needed.

  The prior distribution can be written in the exponential family form where the natural parameter is $\eta=(-\alpha-1,-\beta)$ and the sufficient statistic is $T(\x)=(\log \x, \x^{-1})$. The logarithm of the prior distribution is given by
\begin{align}
\log p(\x) \propto -(\alpha+1)\log \x -\beta \x^{-1}.
\end{align}
In order to compute the posterior using the proposed linearization approach, first  the log-likelihood function is computed
\begin{align}
-2 \mathcal{L}(\x,\y)\pluseq &\log (\x+\sigma^2) + \frac{\y^2}{\x+\sigma^2} \label{eq:llexnormaligammaunsplit}.
\end{align} 
The linearization of the  log-likelihood with respect to the sufficient statistic of the prior distribution can be carried out in several ways, four of which are listed in the following;

\vspace{3mm}
\textit{Solution 1:}{ We can linearize the right-hand-side of \eqref{eq:llexnormaligammaunsplit} with respect to $\log \x$ using Lemma~\ref{lem:invertible-t} by letting $\z \triangleq \log \x$ and subsequent linearization with respect to $\z$ about $\widehat{\z}\triangleq \log{\widehat{\x}}$ which gives
\begin{align}
-2 \mathcal{L}(\x,\y)
\plusapprox 
  \left(\frac{\widehat{\x}}{\widehat{\x}+\sigma^2}+\frac{-\y^2 \widehat{\x}}{(\widehat{\x}+\sigma^2)^2}\right)\log \x,
\end{align}  
where $\plusapprox$ means approximation up to an additive constant with respect to all sufficient statistics (\ie the elements of $T(\x)$). The approximate likelihood can be found as
\begin{align}
\widehat{p_1}(\y|\x)\propto \x^{-\frac{\widehat{\x}}{2(\widehat{\x}+\sigma^2)^2}(\widehat{\x}+\sigma^2-\y^2)}
\end{align}
which is not integrable with respect to $\y$ since $\widehat{\x}>0$ and $\widehat{p_1}(\y|\x)$ goes to infinity as $\y$ goes to infinity. }

\vspace{3mm}
\textit{Solution 2:}{ We can linearize the right-hand-side of \eqref{eq:llexnormaligammaunsplit} with respect to $\x^{-1}$ by letting $\z \triangleq \x^{-1}$ and subsequent linearization with respect to $\z$ about $\widehat{\z}\triangleq {\widehat{\x}^{-1}}$ using  Lemma~\ref{lem:invertible-t} which gives
\begin{align}
-2 \mathcal{L}(\x,\y)
\plusapprox
  \left(\frac{-\widehat{\x}^{2}}{\widehat{\x}+\sigma^2}+\frac{\y^2 \widehat{\x}^{2}}{(\widehat{\x}+\sigma^2)^2}\right)\x^{-1}.
\end{align} 
The approximate likelihood can be found as
\begin{align}
\widehat{p_2}(\y|\x)\propto \exp\left({-\frac{\widehat{\x}^2}{2(\widehat{\x}+\sigma^2)^2 \x}(\y^2-\widehat{\x}-\sigma^2)}\right),
\end{align}
which is integrable with respect to $\y$. However, the approximate posterior obtains the form
\begin{align}
&\widehat{p}(\x|\y)\nonumber\\
&\propto \exp\left(-(\alpha+1)\log \x -\left({\frac{\widehat{\x}^2(\y^2-\widehat{\x}-\sigma^2)}{2(\widehat{\x}+\sigma^2)^2 }}+\beta\right)\x^{-1}\right),
\end{align}}
which is not integrable with respect to $\x$ for a small $\y$ such that ${\frac{\widehat{\x}^2(\y^2-\widehat{\x}-\sigma^2)}{2(\widehat{\x}+\sigma^2)^2 }}+\beta<0$.

\vspace{3mm}
\textit{Solution 3:}{ We can linearize the first term  on the right-hand-side of \eqref{eq:llexnormaligammaunsplit} with respect to $\log \x$ and the second term with respect to $\x^{-1}$ about $\log \widehat{\x}$ and ${\widehat{\x}^{-1}}$, respectively, using  Lemma~\ref{lem:invertible-t} which gives
\begin{align}
-2 \mathcal{L}(\x,\y)\plusapprox & \frac{\widehat{\x}}{\widehat{\x}+\sigma^2}\log\x + \frac{\y^2 \widehat{\x}^{2}}{(\widehat{\x}+\sigma^2)^2}\x^{-1}.
\end{align} 
The approximate likelihood can be found as
\begin{align}
\widehat{p_3}(\y|\x)\propto \x^{-\frac{\widehat{\x}}{2(\widehat{\x}+\sigma^2)}} \exp\left( -\frac{\y^2 \widehat{\x}^{2}}{2\x(\widehat{\x}+\sigma^2)^2}\right),
\end{align}
which is integrable with respect to $\y$.  The approximate posterior remains integrable with respect to $\x$ for $\widehat{\x}>0$.  }

\vspace{3mm}
\textit{Solution 4:}{ We can first expand  the log-likelihood as in
\begin{align}
-2 \mathcal{L}(\x,\y)\pluseq &\log \x + \log\left(1+\frac{\sigma^2}{\x}\right) + \frac{\y^2}{\x+\sigma^2} \label{eq:llexnormaligamma}
\end{align} 
and then linearize the  last two terms on the right hand side of \eqref{eq:llexnormaligamma} with respect to  $\z\triangleq\x^{-1}$ about the nominal point $ \widehat{\z}=\widehat{\x}^{-1}$ using Lemma~\ref{lem:invertible-t} which gives
\begin{align}
-2 \mathcal{L}(\x,\y)\plusapprox & \log\x +  \left(\frac{\sigma^2\widehat{\x}}{\sigma^2+\widehat{\x}}+\frac{\y^2\widehat{\x}^2}{(\sigma^2+\widehat{\x})^2}\right)\x^{-1}.
\end{align}  
The approximate likelihood can be found as
\begin{align}
\widehat{p_4}(\y|\x)\propto \x^{-\frac{1}{2}} \exp\left(-\frac{1}{2\x}\left(\frac{\sigma^2\widehat{\x}}{\sigma^2+\widehat{\x}}+\frac{\y^2\widehat{\x}^2}{(\sigma^2+\widehat{\x})^2}\right)\right)
\end{align}
which is integrable with respect to $\y$. The posterior remains integrable with respect to $\x$ for $\widehat{\x}>0$.  }

Integrability of the approximate posterior is not guaranteed for the first two solutions, while the last two solutions will result in an  inverse gamma approximate posterior distribution. 
The parameters of the posterior depends on the linearization point $\widehat{\x}$ among other factors. A candidate point for linearization is the prior mean  $\widehat{\x}=\frac{\alpha}{\beta}$. 
\hfill$\blacksquare$
\end{exmp}


As pointed out in Theorem~\ref{theorem:iffconjugatelikelihood} and illustrated in Example~\ref{ex:normal-igamma},  linearity of the log-likelihood with respect to sufficient statistic of the prior is a necessary condition for conjugacy. 
Hence, even after succeeding in linearization of the log-likelihood with respect to $T(\x)$, the integrability of the posterior distribution should be examined. Consequently, the linearization should be done skillfully such that the approximation error is minimized and the approximate posterior remains integrable.


In the rest of this paper we will focus on exemplifying the application of the linearization technique in a specific Bayesian inference problem in Section~\ref{sec:ettviaLLL} and the related numerical simulations in Section~\ref{sec:numsim}. In this specific example, two of the sufficient statistics is not invertible as it is required in Lemma~\ref{lem:invertible-t}. Hence, a remedy is suggested and used in the example.

\section{Extended Target Tracking} \label{sec:ettviaLLL}

\subsection{The problem formulation}
\label{sec:problemformulation}
In the Bayesian extended target tracking (ETT) framework~\cite{Koch2008} the state of an extended target consists of a kinematic state  and a symmetric positive definite matrix representing the extent state.
Suppose that the extended target's kinematic state and the target's extent state are denoted by  $x_k$ and $X_k$, respectively. Further, we assume that the measurements $\mathcal{Y}_k\triangleq\{ y_k^j\in \mathbb{R}^d \}_{j=1}^{m_k}$ generated by the extended target are independent and identically distributed (conditioned on  $x_k$ and $X_k$) as  $y_k^j\thicksim\N(y_k^j;Hx_k,sX_k+R)$, see Fig.~\ref{fig:ETTbayesnet}. The following measurement likelihood which will be used in this paper was first introduced in~\cite{FeldmanFK2011}
\begin{align}
p(\mathcal{Y}_k|x_k,X_k)=\prod_{j=1}^{m_k}\N(y_k^j;Hx_k,sX_k+R),\label{eqn:likelihood}
\end{align}
where
\begin{itemize}
\item $m_k$ is the number of measurements at time $k$.
\item $s$ is a real positive scalar constant.
\item $R$ is the measurement noise covariance.
\end{itemize}
We assume the following prior form on the kinematic state and target extent state.
\begin{equation}
\label{eqn:prior}
\begin{split}
p(&x_k,X_k|\mathcal{Y}_{0:k-1})\\
&=\N(x_k;x_{k|k-1},P_{k|k-1})\IW(X_k;\nu_{k|k-1},V_{k|k-1}) 
\end{split}
\end{equation}
where $x_k\in\mathbb{R}^n$, $X_k\in\mathbb{R}^{d\times d}_{>0}$ and the double time index  ``$a|b$" is read  ``at time  $a$ using measurements up to and including time $b$".

The inverse Wishart distribution $\IW(X;\nu,V)$ we use in this work is given in the following form.
\begin{equation}
\IW(X;\nu,V)\triangleq \frac{|V|^{\frac{1}{2}(\nu-d-1)}\exp\tr\left(-\frac{1}{2}VX^{-1}\right)}{2^{\frac{1}{2}(\nu-d-1)d}\Gamma_d\left[\frac{1}{2}(\nu-d-1)\right]|X|^{\frac{\nu}{2}}}
\end{equation}
where $X$ is a symmetric positive definite matrix of dimension $d\times d$, $\nu>2d$ is the scalar degrees of freedom and $V$ is a symmetric positive definite matrix of dimension $d\times d$ and is called the scale matrix. This form of the inverse Wishart distribution is the same one used in the well-known reference~\cite{GuptaN:2000}.

No analytical  solution for the posterior exists. The reason for the lack of an analytical solution for the posterior corresponding to the likelihood~\eqref{eqn:likelihood} and the prior~\eqref{eqn:prior} is both the covariance addition in the Gaussian distributions on the right hand side of~\eqref{eqn:likelihood} and the intertwined nature of the kinematic state and extent state in the likelihood. In order to get rid of the covariance addition, latent variables are defined in~\cite{Orguner2012} and the problem of intertwined kinematic and extent states are solved using variational approximation. 
In \cite{FeldmanFK2011} the authors design an unbiased estimator. Both of these measurement updates can be used in a recursive estimation framework which requires the posterior to be in the same form as the prior as in 
\begin{equation}
\label{eqn:posterior}
\begin{split}
p(x_k,X_k|\mathcal{Y}_{0:k})\approx \N(x_k;x_{k|k},P_{k|k})\IW(X_k;\nu_{k|k},V_{k|k}).
\end{split}
\end{equation}
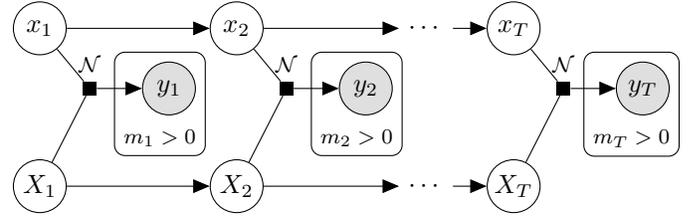
\begin{figure}[t]
  \begin{center}
    \begin{tikzpicture}
        \node[obs] (y1) {$y_1$} ; %
				\plate {} {(y1)} {$m_1 >0$};
        \node[latent, left=of y1, yshift=0.8cm] (x1) {$x_1$} ; %
        \node[latent, left=of y1, yshift=-1.3cm] (X1) {$X_1$} ; %
        \factor[left=of y1, xshift=-0.2cm] {y-factor} {$\mathcal{N}$} {} {};
        \factoredge {x1,X1} {y-factor} {y1} ; %
				
				\node[obs,right=of y1, xshift=9mm] (y2) {$y_2$} ; %
				\plate {} {(y2)} {$m_2 >0$};
        \node[latent, left=of y2, yshift=0.8cm] (x2) {$x_2$} ; %
        \node[latent, left=of y2, yshift=-1.3cm] (X2) {$X_2$} ; %
        \factor[left=of y2, xshift=-0.2cm] {y-factor} {$\mathcal{N}$} {} {};
        \factoredge {x2,X2} {y-factor} {y2} ; %
				
				\node[right=of y2, xshift=8mm] (y3) {$$} ; %
        \node[ left=of y3, yshift=0.8cm] (x3) {$\cdots$} ; %
        \node[ left=of y3, yshift=-1.3cm] (X3) {$\cdots$} ; %

				\node[obs,right=of y4, xshift=-1mm] (yT) {$y_T$} ; %
				\plate {} {(yT)} {$m_T >0$};
        \node[latent, left=of yT, yshift=0.8cm] (xT) {$x_T$} ; %
        \node[latent, left=of yT, yshift=-1.3cm] (XT) {$X_T$} ; %
        \factor[left=of yT, xshift=-0.2cm] {y-factor} {$\mathcal{N}$} {} {};
        \factoredge {xT,XT} {y-factor} {yT} ; %
				
  \edge {x1} {x2} ; %
  \edge {X1}  {X2} ; %
	\edge {x2} {x3} ; %
  \edge {X2}  {X3} ; %
	\edge {x3} {xT} ; %
  \edge {X3}  {XT} ; %
\end{tikzpicture}
  \end{center}
  \caption{A probabilistic graphical model for extended targets with kinematic state $x_k$, extent state $X_k$ and measurements $y_k$.}
	\label{fig:ETTbayesnet}
	\vspace{-5mm}
\end{figure}
\subsection{Solution proposed by Feldmann et al.~\cite{FeldmanFK2011}}
In~\cite{FeldmanFK2011} the authors cleverly design a measurement update based on unbiasedness properties for the measurement model~\eqref{eqn:likelihood} to calculate $x_{k|k}$, $P_{k|k}$, $\nu_{k|k}$ and $V_{k|k}$ in the posterior \eqref{eqn:posterior}. 

The kinematic state $x_k$ is updated as follows:
\begin{align}
x_{k|k}&=x_{k|k-1}+K_k(\bar{y}_k-Hx_{k|k-1})\\
P_{k|k}&=P_{k|k-1}-K_kS_kK_k^{\t{}}
\end{align}
where
\begin{align}
K_k&=P_{k|k-1}H^{\t{}}S_k^{-1}\\
S_k&=HP_{k|k-1}H^{\t{}}+\frac{1}{m_k}(sX_{k|k-1}+R)\\
X_{k|k-1}&=\frac{V_{k|k-1}}{\nu_{k|k-1}-2d-2}
\end{align}
and
\begin{align}
\bar{y}_k\triangleq \frac{1}{m_k}\sum_{j=1}^{m_k} y_k^j
\end{align}
is the mean of the measurements at time $k$.

The extent state updates given in~\cite{FeldmanFK2011} is in the following form
\begin{align}
\nu_{k|k}=&\nu_{k|k-1}+m_k\\
V_{k|k}=&V_{k|k-1}+M_k^{\FFK} \label{eq:extentupdate}
\end{align}
where $M_k^{\FFK}$ is a given positive definite update matrix and the superscript $\cdot^{\FFK}$ which is used to ease presentation consists of the authors initials in~\cite{FeldmanFK2011}. Before describing the construction of the matrix $M_k^{\FFK}$ of~\cite{FeldmanFK2011} let us define  the measurement spread $Y_k$ from the predicted measurement $H x_{k|k-1}$ as follows.
\begin{align}
Y_k\triangleq\frac{1}{m_k}\sum_{j=1}^{m_k}(y_k^j-H x_{k|k-1})(y_k^j-H x_{k|k-1})^\t
\end{align}
Feldmann et al. first write the measurement spread $Y_k$ as  
\begin{align}
Y_k=Y_k^1+Y_k^2.\label{eqn:decompositionYk}
\end{align}
The summands $Y_k^1$ and $Y_k^2$ are defined as
\begin{align}
Y_k^1\triangleq& (\bar{y}_k-H x_{k|k-1})(\bar{y}_k-H x_{k|k-1})^\t,\\
Y_k^2\triangleq& \frac{1}{m_k}\sum_{j=1}^{m_k} (y_k^j-\bar{y}_k) (y_k^j-\bar{y}_k)^\t{}.
\end{align}
When we take the conditional expected values of $Y_k^1$ and $Y_k^2$, we obtain
\begin{align}
\overline{Y}_k^1\triangleq& E\left[\left.Y_k^1\right|X_k=X_{k|k-1}\right]\nonumber\\
&=HP_{k|k-1}H^\t{} +\frac{sX_{k|k-1}+R}{m_k},\\
\overline{Y}_k^2\triangleq& E\left[\left.Y_k^2\right|X_k=X_{k|k-1}\right]\nonumber\\
&=\frac{m_k-1}{m_k}(sX_{k|k-1}+R).
\end{align}
The matrix $M_k^{\FFK}$ is then proposed to be
\begin{align}
\begin{split}
M&_k^{\FFK}=X_{k|k-1}^{1/2}(\overline{Y}_k^1)^{-1/2}Y_k^1(\overline{Y}_k^1)^{-1/2}X_{k|k-1}^{1/2}\\
&+(m_k-1)X_{k|k-1}^{1/2}(\overline{Y}_k^2)^{-1/2}Y_k^2(\overline{Y}_k^2)^{-1/2}X_{k|k-1}^{1/2}.
\end{split}
\end{align}
The conditional expected value of $M_k^{\FFK}$ is
\begin{align}
E\left[\left.M_k^{\FFK}\right|X_k=X_{k|k-1}\right]=m_kX_{k|k-1},
\end{align}
which results in an unbiased estimator. 
\subsection{ETT via log-likelihood linearization} \label{sec:ettLLL}
A new measurement update is obtained by performing linearization of the logarithm of the likelihood function~\eqref{eqn:likelihood} with respect to sufficient statistic of the prior distribution \eqref{eqn:prior}. The sufficient statistics of the prior distribution~\eqref{eqn:prior} are given as follows.
\begin{align}
\label{eq:sufficientstatsETT}
T(x_k,X_k)=(x_k,x_kx_k^\t,X_k^{-1},\log|X_k|)
\end{align}
As seen above the second and the fourth elements of $T(\cdot)$ are not invertible. In the following lemma, using log-likelihood linearization (LLL) via a first order Taylor series expansion, we approximate the likelihood function~\eqref{eqn:likelihood} to obtain a conjugate likelihood (with respect to the prior ~\eqref{eqn:prior}) with sufficient statistics given in \eqref{eq:sufficientstatsETT}.


\begin{lemma}\label{lem:ETTlinearization}The likelihood $\prod_{j=1}^m\N(y^j;Hx,sX+R)$ can be approximated up to a multiplicative constant\footnote{i.e., constant with respect to the variables $X$ and $x$.} by a first order Taylor series approximation around the nominal points $\widehat{X}$ (for variable $X$) and $\hat{x}$ (for variable $x$) as
\begin{align}
\prod_{j=1}^m\N&(y^j;Hx,sX+R)\nonumber\\
&\approx \left(\prod_{j=1}^m \N(y^j;Hx,s\widehat{X}+R)\right)\IW\left(X;m,M\right)\label{eq:lemrighthandside}
\end{align}
where
\begin{equation}
\label{eqn:Malternative}
\begin{split}
M=&m\widehat{X}+ms\widehat{X}(s\widehat{X}+R)^{-1}\\
&\times\left[Y_k-(s\widehat{X}+R)\right] (s\widehat{X}+R)^{-1}\widehat{X}.
\end{split}
\end{equation}
\begin{proof}Proof is given in Appendix~\ref{sec:appendixproof} for the sake of clarity.
\end{proof}
\end{lemma}
Lemma \ref{lem:ETTlinearization} states that the likelihood~\eqref{eqn:likelihood} can be approximately factorized into two independent likelihood terms corresponding to kinematic and extent states.
This type of factorization gives independent measurement updates for the kinematic state and the extent state. For this purpose, one can set $\widehat{X}=\frac{V_{k|k-1}}{\nu_{k|k-1}-2d-2}$ and $\hat{x}=x_{k|k-1}$ in order to find the factors of Lemma~\ref{lem:ETTlinearization}. Using the conjugate likelihood factors in~\eqref{eq:lemrighthandside}, the posterior density $p(x_k,X_k||\mathcal{Y}_{0:k})$ is given in the form of \eqref{eqn:posterior}   with the update parameters given below.  
\begin{align}
&x_{k|k}=P_{k|k} \nonumber\\
&\times\left(P_{k|k-1}^{-1}x_{k|k-1}+m_kH^\t (sX_{k|k-1}+R)^{-1}\bar{y}_k\right),\\
&P_{k|k}=\left(P_{k|k-1}^{-1}+m_kH^\t (sX_{k|k-1}+R)^{-1}H\right)^{-1},
\end{align}
where
\begin{align}
X_{k|k-1}=&\frac{V_{k|k-1}}{\nu_{k|k-1}-2d-2},\\
\bar{y}_k=&\frac{1}{m_k}\sum_{j=1}^{m_k} y_k^j.
\end{align}
Note that the kinematic updates are the same as the kinematic updates proposed in~\cite{FeldmanFK2011}.

The extent state is updated as given in  
\begin{align}
\nu_{k|k}=&\nu_{k|k-1}+m_k\\
V_{k|k}=&V_{k|k-1}+M_k^{\LLL}
\end{align}
where
\begin{equation}
\begin{split}
M&_k^{\LLL}=m_kX_{k|k-1}+m_ksX_{k|k-1}(sX_{k|k-1}+R)^{-1}\\
&\times\left[Y_k-(sX_{k|k-1}+R)\right] (sX_{k|k-1}+R)^{-1}X_{k|k-1}.
\end{split}
\end{equation}
This form of the update suggests that the matrix $Y_k$ serves as a pseudo-measurement for the quantity $sX_k+R$. If $Y_k$ is larger than the current predicted estimate of $sX_k+R$  (i.e., $sX_{k|k-1}+R$) then a positive matrix quantity is added to the statistics $V_{k|k-1}$ and vice versa.

We here note that 
\begin{align}
\E\left[\left.Y_k\right|X_k=X_{k|k-1}\right]=HP_{k|k-1}H^\t{}+sX_{k|k-1}+R
\end{align}
where the expectation is taken over all the measurements at time $k$ and the estimate $x_{k|k-1}$. Hence we conclude that expected value of the second term on the right hand side of~\eqref{eqn:Malternative} is always positive which makes the current update biased. Another drawback of the update is that the uncertainty of the predicted estimate $x_{k|k-1}$ does not affect the update. In order to solve both problems we modify the quantity $M_k^{\LLL}$ as follows.
\begin{align}
M_k^{\ULL}\triangleq& m_kX_{k|k-1}\nonumber\\
&+m_ksX_{k|k-1}(HP_{k|k-1}H^\t{}+sX_{k|k-1}+R)^{-1}\nonumber\\
&\times\left[Y_k-(HP_{k|k-1}H^\t{}+sX_{k|k-1}+R)\right]\nonumber\\
&\times(HP_{k|k-1}H^\t{}+sX_{k|k-1}+R)^{-1}X_{k|k-1}.\label{eqn:Mmodified}
\end{align}
Note that in the current update the difference 
\begin{align}
Y_k-(HP_{k|k-1}H^\t{}+sX_{k|k-1}+R)
\end{align}
has zero conditional mean which makes the update unbiased and the terms $(HP_{k|k-1}H^\t{}+sX_{k|k-1}+R)^{-1}$ decrease with increasing uncertainty in the predicted estimate $x_{k|k-1}$ which makes the update term smaller. In the following we will call the proposed unbiased measurement update based on linearization of the likelihood in the log domain as $\ULL$ for ease of presentation.

\section{Numerical simulation} \label{sec:numsim}
In this section the newly proposed measurement update $\ULL$, is compared with the update based on variational Bayes ~\cite{Orguner2012} and  update proposed by Feldmann et al. in~\cite{FeldmanFK2011} which will be referred to as $\VB$ and $\FFK$, respectively. 
In section~\ref{sec:simMC}, we will use a simulation scenario which is partly based on the simulation scenario described in~\cite[Section IV]{Orguner2012} which in turn is based on~\cite[Section 6.1]{BaumSim}. In section~\ref{sec:simETT} an ETT scenario based on the numerical simulation   described in~\cite[VI-B]{FeldmanFK2011} will be presented.

\subsection{Monte-Carlo simulations} \label{sec:simMC}
A planar extended target in the two dimensional space  whose true kinematic state $x_k^0$ and the extent state $X_k^0$ are 
\begin{subequations}
\label{eq:initialstate}
\begin{align}
x_k^0&=[0\;m,0\;m,100\;m/s,100\;m/s]^\t{}\\
X_k^0&=E_k \diag([300^2\; m^2,200^2\; m^2])E_k^\t{}
\end{align}
\end{subequations}
is considered. Here, $E_k\triangleq[e_1,e_2]$ is a $2\times 2$ matrix whose columns $e_1$ and $e_2$ are the normalized eigenvectors
of $X_k^0$ which are $e_1\triangleq \frac{1}{\sqrt{2}}[1,1]^\t{}$ and $e_2\triangleq \frac{1}{\sqrt{2}}[1,-1]^\t{}$. For the extended target with these true  parameters, we conduct a Monte Carlo
(MC) simulations to quantify the differences between the measurement updates FFK, ULL and VB.
For each MC run  we assume that the initial predicted target density for
all three methods is the same and has the following structure:
\begin{equation}
\begin{split}
p(&x_k,X_k|\mathcal{Y}_{0:k-1})\\
&=\N(x_k;x_{k|k-1}^j,P_{k|k-1} )\IW(X_k;\nu_{k|k-1}^j,V_{k|k-1}^j)
\end{split}
\end{equation}
where superscript $j$ indicates the $j$th MC run. 
The quantities $x_{k|k-1}^j$, $P_{k|k-1}$ for the kinematic state are selected as
\begin{subequations}
\begin{equation}
x_{k|k-1}^j \thicksim \N\left(\cdot;x_k^0,\frac{P_{k|k-1}}{\alpha_k}\right)\label{eq:ak}
\end{equation}
\begin{equation}
P_{k|k-1} = \diag\left([50^2, 50^2, 10^2,10^2]\right)
\end{equation}
\end{subequations}
where the scalar $\alpha_k$ is the scaling parameter for the covariance  of the density in \eqref{eq:ak} to adjust the distribution of $x_{k|k-1}^j$ in the MC simulation study.
The quantities $\nu_{k|k-1}^j$ and $V_{k|k-1}^j$ are selected as
\begin{subequations}
\begin{equation}
\nu_{k|k-1}^j =\max(7,\nu^j \thicksim \Poisson(100) )
\end{equation}
\begin{equation}
\left.\frac{V_{k|k-1}^j}{\nu_{k|k-1}^j-2d-2}\right|\nu_{k|k-1}^j \thicksim \W\left(\cdot;\delta_k,\frac{X_k^0}{\delta_k}\right)
\end{equation}
\end{subequations}
where the functions $\Poisson(\lambda)$ and $\W(\cdot;n,\Psi)$ represent the Poisson
density with expected value $\lambda$ and Wishart density with degrees of freedom $n$ and scale matrix  $\Psi$. The scalar $\delta_k$ controls the variance of the Wishart density in this study. For the $j^{th}$ MC run, $m_k^j$ measurements are generated according to
\begin{subequations}
\label{eq:MCmeasurement}
\begin{align}
m_k^j&=\max(2,m^j\thicksim\Poisson(10)),\\
y_k^{j,i}&\thicksim\N(\cdot;Hx_k^0,sX_k^0+R)
\end{align}
\end{subequations}
 where $s=0.25$, $R=100^2I_2\;m^2$ and $H=[I_2,0_{2\times2}]$. Matrices $I_2$ and $0_{2\times2}$ are the identity matrix and zero matrix of size $2\times 2$, respectively. 
The performance of the measurement updates are compared for various levels of accuracy of the initial predicted target density in view of the  optimal Bayesian solution. To this end, the optimal solution is numerically computed via
importance sampling using the predicted density as the proposal density. Total of $10^5$ samples are generated from the predicted density and  the normalized importance weights are calculated from the likelihood function given in \eqref{eqn:likelihood}. Using the importance weights, the posterior expected values of the kinematic state $x_k$ and extent state $X_k$ are calculated.  The resulting expected values  are referred to as $x_{k|k}^{opt}$ and $X_{k|k}^{opt}$ respectively.

To change the level of accuracy for the predicted target density, the parameters $\alpha_k$ are selected on a linear grid of $40$ values between $1$ and $50$ while values of $\delta_k$ were selected on a logarithmic grid of $40$ values between $2$ and $1000$. 
For each pair of $\alpha_k$
 and $\delta_k$ out of the 40 pairs, $N_{MC}=1000$ runs have been made by selecting the parameters $x_{k|k-1}^j$, $P_{k|k-1}$, $\nu_{k|k-1}^j$, $V_{k|k-1}^j$ and $\mathcal{Y}_{k}^j$ as above and the estimates 	 $x_{k|k}^{U,j}$ and $X_{k|k}^{U,j}$ for $U \in\{\FFK,\VB,\ULL\}$ are calculated\footnote{In each MC run  20 iterations were performed for the VB solution, although the VB solution would usually converge within the first 5 iterates.}.

Using the results of the MC runs, the average error metrics $E_x^U$ and $E_X^U$ for $U \in\{\FFK,\VB,\ULL\}$ 
for the kinematic state and the extent state, respectively, are calculated as follows:
\begin{align}
E_x^U &\triangleq \left ( \frac{1}{d   N_{MC}} \sum_{j=1}^{N_{MC}} {\left\| H(x_{k|k}^{U,j}-x_{k|k}^{opt,j}) \right\|_2^2} \right)^{\frac{1}{2}},\\ 
E_X^U &\triangleq \left ( \frac{1}{d^2 N_{MC}} \sum_{j=1}^{N_{MC}} {\tr (X_{k|k}^{U,j}-X_{k|k}^{opt,j})^2} \right)^{\frac{1}{4}}.
\end{align}

\begin{figure}
\centering
\includegraphics[width=.45\textwidth]{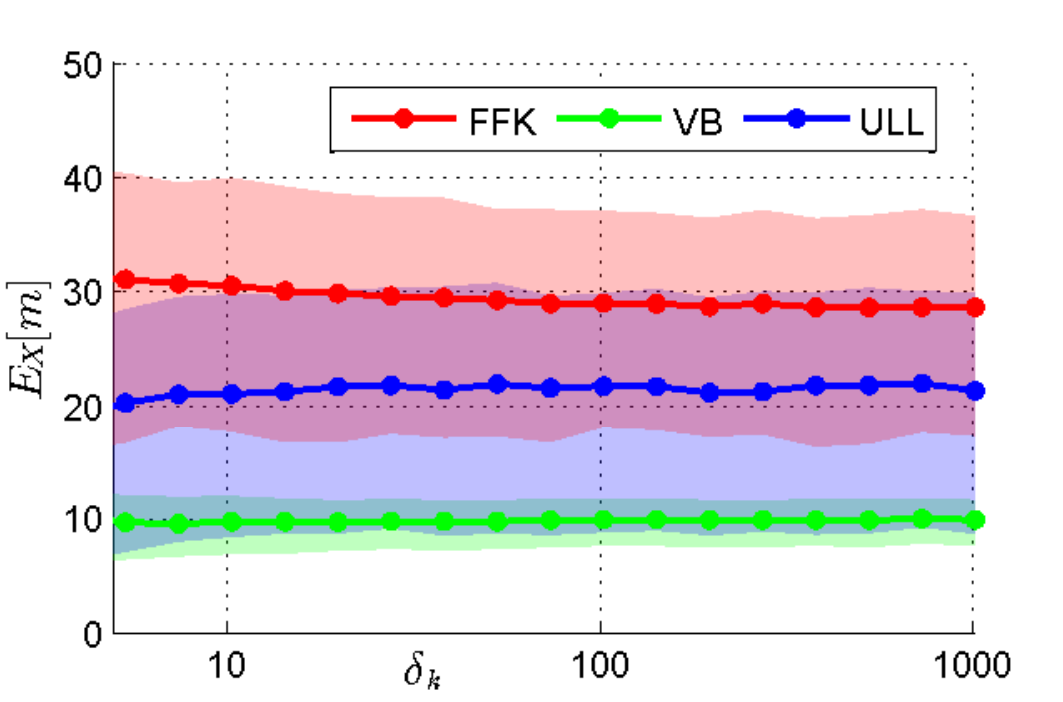}			
\caption{Extent estimation error for FFK, VB and ULL with respect to the optimal Bayesian solution. Solid lines show the mean error $E_X$ and shaded areas of respective colors illustrate the interval between the fifth and the ninety fifth percentiles.}
\label{fig:EX}	
\end{figure}

\begin{figure}
\centering
\includegraphics[width=.45\textwidth]{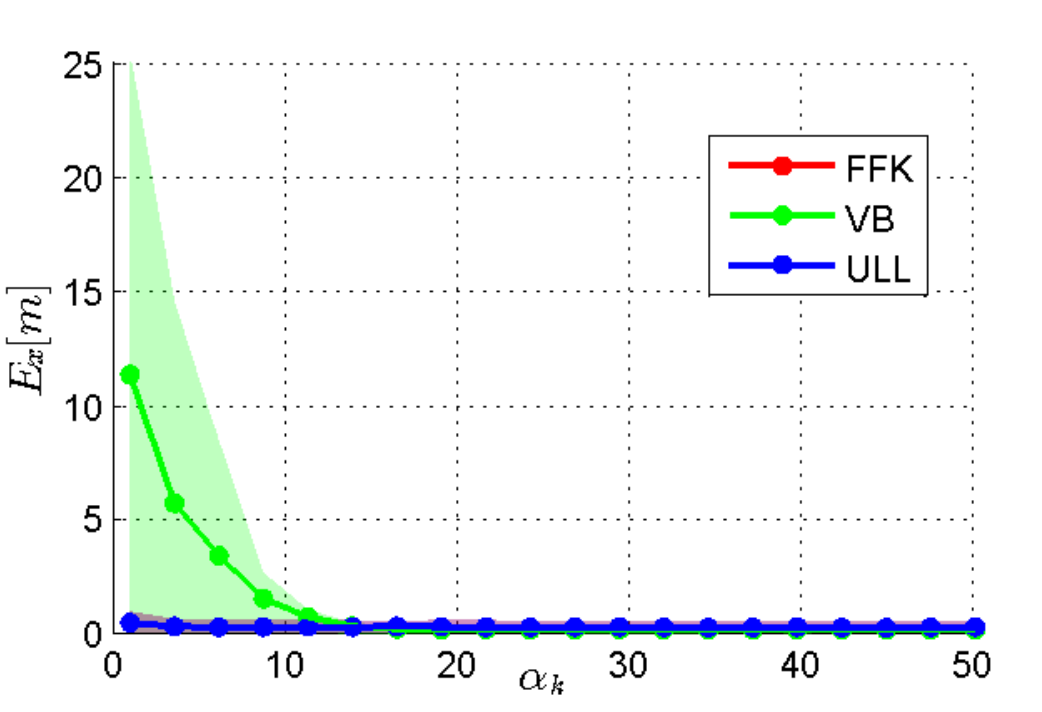}			
\caption{Kinematic state estimation error for FFK, VB and ULL with respect to the optimal Bayesian solution. Solid lines show the mean error $E_x$ and shaded areas of respective colors illustrate the interval between the fifth and the ninety fifth percentiles. Since FFK and ULL use identical kinematic state update (for identical $X_{k|k-1}$)  their respective graphs coincide.}
\label{fig:rms}	
\end{figure}

The errors are illustrated in Fig.~\ref{fig:EX} and Fig.~\ref{fig:rms}.	Since FFK and ULL use identical kinematic state update, their respective graphs coincide. VB has the lowest extent state estimation error. ULL has a lower estimation compared to FFK in the simulations scenario presented here. When the simulation is repeated with $R=50^2 I_2$ FFK performs better than ULL while VB maintains its superior estimation performance in terms of $E_X$.

\subsection{Single extended target tracking scenario}\label{sec:simETT}
\begin{figure}[t]
\centering
\includegraphics[width=.5\textwidth]{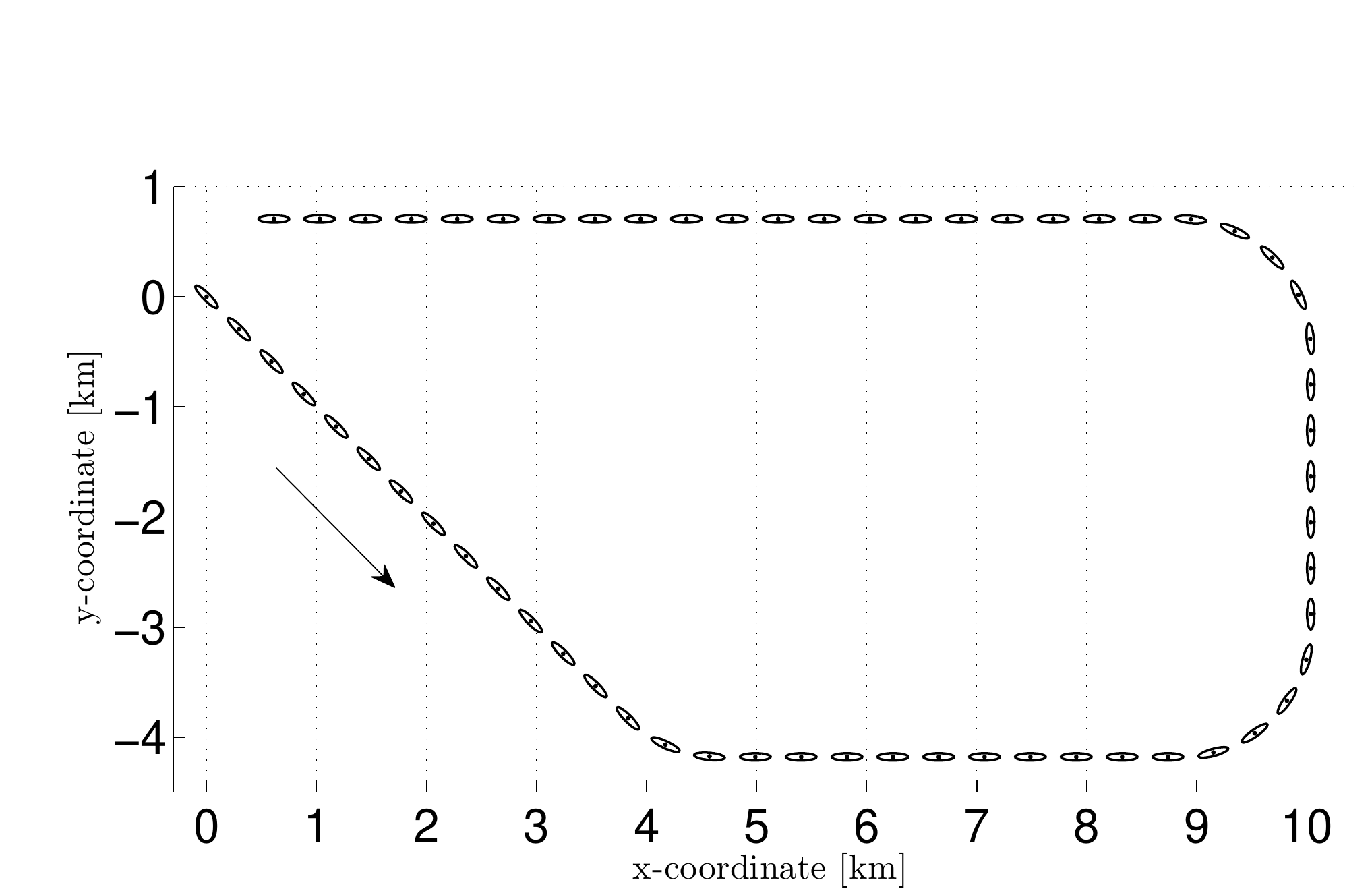}			
\caption{Trajectory of an extended target. For every third scan, the true extended target and its center are shown. }
\label{fig:ETT-trajectory}	
\end{figure}

The performance of measurement update is evaluated in a single extended target tracking scenario where an extended target follows a trajectory in a  two dimensional space illustrated in Fig.\ref{fig:ETT-trajectory}. 
In this simulation scenario, an elliptical extended target with diameters $340$ and $80$ meters moves with a constant velocity of about $50\ \text{km/h}$. 
The extended target's true initial  kinematic state $x_1^0$ and true initial extent state $X_1^0$ are  as follows:
\begin{subequations}
\label{eq:initialstateETT}
\begin{align}
x_1^0&=[0m,0m,9.8m/s,-9.8m/s]^\t{}\\
X_1^0&=E_k \diag([170^2\ m^2,400^2\ m^2])E_k^\t{}
\end{align}
\end{subequations}
where $E_k$ is the $2\times 2$ matrix whose columns are the normalized eigenvectors
of $X_1^0$ which are $e_1\triangleq \frac{1}{\sqrt{2}}[-1,1]^\t{}$ and $e_2\triangleq \frac{1}{\sqrt{2}}[1,1]^\t{}$, \ie $E_k\triangleq[e_1,e_2]$.
\begin{table}[t]
\caption{Comparison of measurement updates for ETT with respect to cycle time and estimation errors. The data are obtained from the single ETT simulation scenario. Since the VB update diverged several time, the errors exceeding $24m$ are set to $24m$ to make the quantitative comparison meaningful.  }
\centering
\begin{tabular}{l c c c}
\toprule
   & \textbf{$E_x[m]$} & \textbf{$E_{X}[m]$}& {\textbf{Cycle times [s]}} \\
\cmidrule(lr){2-2} \cmidrule(lr){3-3} \cmidrule(lr){4-4}
\textbf{Update } & Mean $\pm$ St.Dev.       & Mean $\pm$ St.Dev.    &  Mean$\pm$ St.Dev. \\
\midrule
FFK    & 15.4581  $\pm$  0.9943   & 19.4354   $\pm$ 0.6894   & 0.1627   $\pm$ 0.0000        \\
VB     &    16.3447  $\pm$  1.1913   & 19.8340  $\pm$  0.7764 &   3.2547  $\pm$  0.0002       \\
ULL    & 15.5204  $\pm$  0.9685  & 19.2356  $\pm$  0.6680   & 0.1293 $\pm$   0.0000       \\
\bottomrule
\end{tabular}
\label{table:cycle_times}
\end{table}

\begin{figure}[t]
\centering
\includegraphics[width=.45\textwidth]{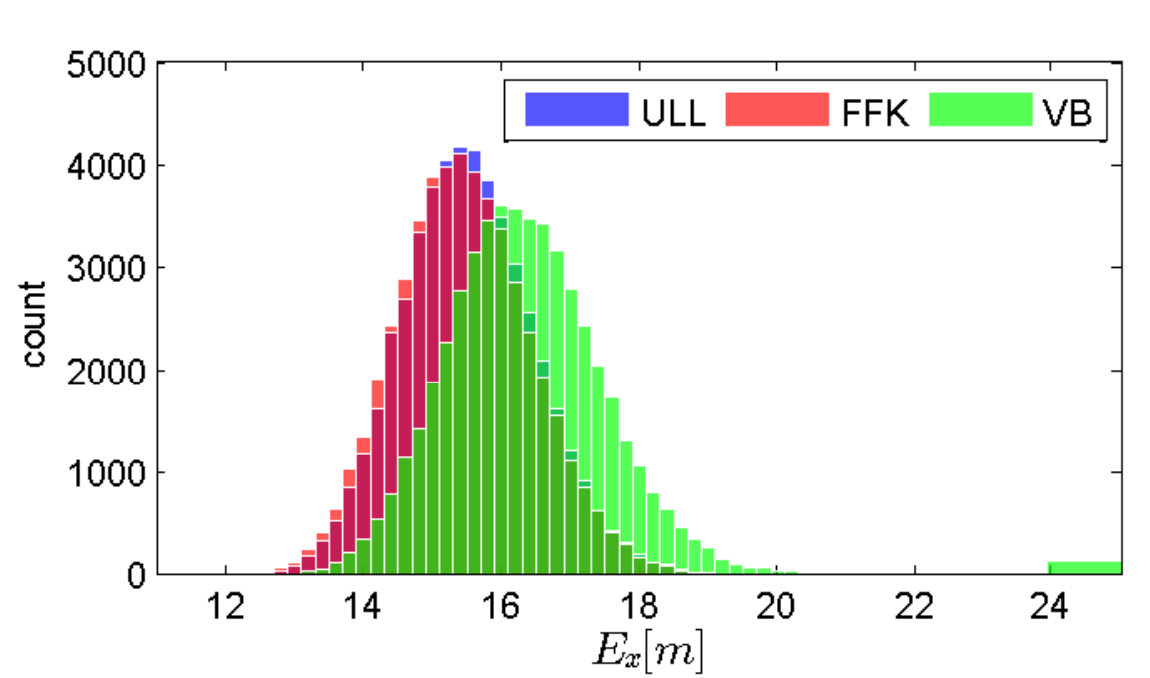}			
\caption{Histogram of the  root mean square error of the kinematic state $E_{x}^{U,j}$ for $U \in\{\FFK,\VB,\ULL\}$ in a single ETT scenario.}
\label{fig:ETT_rmse}	
\end{figure}

\begin{figure}[t]
\centering
\includegraphics[width=.45\textwidth]{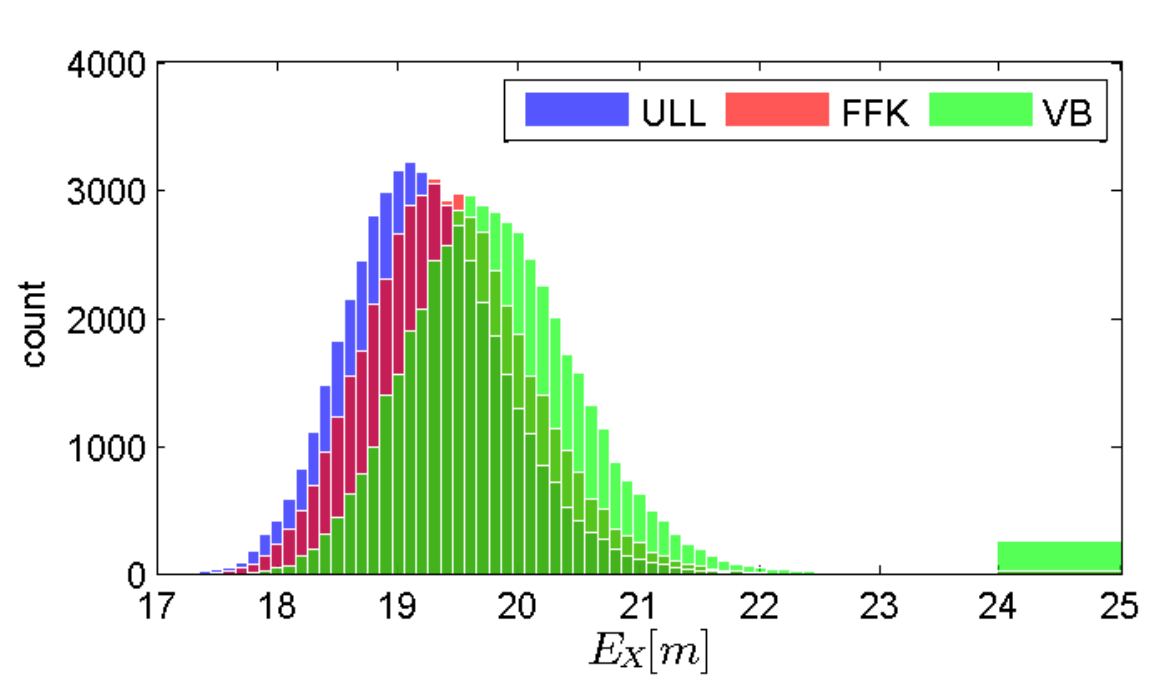}			
\caption{Histogram of the  average extent state estimation error $E_{X}^{U,j}$ for $U \in\{\FFK,\VB,\ULL\}$ in a single ETT scenario.}
\label{fig:ETT_Xnormalized}	
\end{figure}
We conduct MC study to quantify the difference between different measurement updates in a single extended target tracking scenario.
The measurement set $\mathcal{Y}_{k}^j=\{y_k^{j,i}\}_{i=1}^{m_k^j}$ for the $j^{th}$ MC run is generated according to~\eqref{eq:MCmeasurement}
 where $s=0.25$, $R=20^2I_2m^2$ and $H=[I_2,0_{2\times 2}]$ and $1\leq k\leq 181$.

In the  filters, the kinematic state vector consists of the position and the velocity $x_k=[p_x,p_y,\dot{p}_x,\dot{p}_y]^{\t{}}$, which evolves according to  the discrete-time constant velocity model  in two dimensional Cartesian coordinates \cite{Rongli2003} where $p(x_{k+1}|x_k)= \N(x_{k+1};A_kx_k,Q_k)$ and
\begin{align*}
				 &A_k= \begin{bmatrix} I_2 & \tau I_2\\ 0_2 & I_2 \end{bmatrix},   & Q_k= \sigma_\nu^2\begin{bmatrix} \frac{\tau^4}{4}I_2 & \frac{\tau^3}{2} I_2\\ \frac{\tau^3}{2} I_2& \tau^2I_2 \end{bmatrix},	\\
				 &\tau=10s,  & \sigma_v =0.1m/s^2.
\end{align*} 
In the filters the initial prior  density for the kinematic state and extent state are  
\begin{equation}
\begin{split}
p(x_1,X_1)=\N(x_0;x_{1|0}^j,P_{1|0} )\IW(X_0;\nu_{1|0}^j,V_{1|0}^j)
\end{split}
\end{equation}
where
\begin{subequations}
\begin{align}
&x_{1|0}^j\thicksim \N(\cdot;x_0^0,\frac{P_0}{\alpha_0})  \\
&P_{1|0}=\diag([50^2, 50^2, 10^2,10^2])  \\
&\nu_{1|0}^j=\max(7,\nu^j \thicksim \Poisson(10) )\\
&\frac{V_{1|0}^j}{\nu_{1|0}^j-2d-2}|\nu_{1|0}^j \thicksim \W(\cdot;\delta_0,\frac{X_0^0}{\delta_0})
\end{align}
\end{subequations}
and where $\alpha_0=10$ and $\delta_0=5$.
\subsubsection{Time update}
The expression for the time update of the target's kinematic state in the filter follows the standard Kalman filter time update equations,
\begin{align}
&x_{k+1|k}^{j}=A_k x_{k|k}^{j}\\
&P_{k+1|k}^{j}=A_k P_{k|k}^{j}A_k^\t+Q_k.
\end{align}
The extent state is assumed to be slowly varying. Hence, exponential forgetting strategy \cite{Kulhavar93,Koch2008,caravalho2007} can be used to account for possible changes in the parameters in time. 
In \cite{Ozkan12} it is shown that using the  exponential forgetting factor will produce maximum entropy distribution in the time update for the processes which are slowly varying with unknown dynamics but bounded by a Kullback-Leibler  divergence (KLD) constraint. Forgetting factor is applied to the parameters of the inverse Wishart distribution as follows.
\begin{align}
&\nu_{k+1|k}^{j}=\exp(-{\tau}/{\tau_0}) \nu_{k|k}^{j},\\
&V_{k+1|k}^{j}=\frac{\nu_{k+1|k}^{j}-2d-2}{\nu_{k|k}^{j}-2d-2} V_{k|k}^{j}.
\end{align}
The exponential decay time constant was selected $\tau_0=15$. 
\subsubsection{Evaluation of filters}
$50\: 000$ MC runs are performed where the  parameters $x_{1|0}^j$, $P_{1|0}$, $\nu_{1|0}^j$, $V_{1|0}^j$ and $\mathcal{Y}_{k}^j$ for $1\leq k \leq 181$ are selected as above.
 Using the results of the MC runs, the average error metrics $E_x^{U,j}$ and $E_X^{U,j}$ for $U \in\{\FFK,\VB,\ULL\}$ and $j^{th}$ MC run
for the target's kinematic state and the extent state, respectively, are calculated as follows:
\begin{subequations}
\label{eq:errorsETT}
\begin{align}
E_x^{U,j} &\triangleq \left ( \frac{1}{dK} \sum_{k=1}^{K} {\left\| H(x_{k|k}^{M,j}-x_k) \right\|_2^2} \right)^{\frac{1}{2}},\\ 
E_{X}^{U,j} &\triangleq \left ( \frac{1}{d^2 K} \sum_{k=1}^{K} {\tr (X_{k|k}^{M,j}-X_{k})^2}\right)^{\frac{1}{4}}.
\end{align}
\end{subequations}
The errors (mean$\pm$ standard deviation) and the cycle times (mean$\pm$ standard deviation) for evaluated measurement update rules are given in Table~\ref{table:cycle_times}. The histograms of distribution of the errors defined in \eqref{eq:errorsETT} are given in Fig. \ref{fig:ETT_rmse} and Fig.\ref{fig:ETT_Xnormalized}.

The estimation errors of $\FFK$ and $\ULL$ are comparable but the  proposed update $\ULL$, is less computationally expensive.

\section{conclusion} \label{sec:conclusion}

A Bayesian inference technique based
on Taylor series approximation of the logarithm of the likelihood
function is presented. The proposed approximation is devised for
the case where the prior distribution belongs to the exponential
family of distribution and is continuous.  The logarithm of the
likelihood function is linearized with respect to the sufficient
statistic of the prior distribution in exponential family such that
the posterior obtains the same exponential family form as the
prior. 
Taylor series approximation is used for linearization with respect to the sufficient statistic. However, Taylor series approximation is a  local approximation which is used for approximation of the posterior over the entire support. In spite of this inherent weakness, the proposed algorithm performs well in the numerical simulations that are presented here for extended target tracking. Furthermore, there are numerous successful applications of extended Kalman filter, which is a special case of the proposed algorithm, in theory and practice.

When a Bayesian posterior estimate is needed with limited computational budget which prohibits the use of Monte-Carlo methods or even variational Bayes approximation as in real-time filtering problems, the proposed algorithm offers a  good alternative. 

The comparison of  possible choices for the linearization point and linearization methods with respect to the sufficient statistic are among the future research problems. 
\section{acknowledgments} \label{sec:ack}
The authors would like to thank Henri Nurminen for proof reading and providing comments on an earlier version of the  manuscript. 


\bibliographystyle{IEEEtran}
\bibliography{ieeeabrv,ArdeshiriOG-ABI}
\appendix

\subsection{Proof of Lemma \ref{lem:ETTlinearization}}
\label{sec:appendixproof}
The logarithm of the likelihood $\prod_{j=1}^m\N(y^j;Hx,sX+R)$ is given as
\begin{align}
-&2\sum_{j=1}^m\log\N(y^j;Hx,sX+R)\pluseq m\log |(sX+R)|\nonumber\\
&+\sum_{j=1}^m(y^j-Hx)^\t (sX+R)^{-1} (y^j-Hx)\\
=&m\log |X| +m\log \left|sI+X^{-1/2}RX^{-1/2}\right|\nonumber\\
&+\sum_{j=1}^m(y^j-Hx)^\t (sX+R)^{-1} (y^j-Hx)\\
=&m\log |X| +m\log \left|sI+X^{-1/2}RX^{-1/2}\right|\nonumber\\
&+\sum_{j=1}^m \tr\left[(y^j-Hx)(y^j-Hx)^\t (sX+R)^{-1}\right] \label{eqn:proofloggaussian}
\end{align}
where the sign $\pluseq$ denotes an equality up to an additive constant. We  approximate the right hand side of~\eqref{eqn:proofloggaussian} by a function that is linear in the sufficient statistic 
\begin{align}
T(x,X)=(x,xx^\t,\log|X|,X^{-1}). 
\end{align}
As aforementioned, the statistics $xx^\t$ and $\log|X|$ are not invertible. The remedy we suggest here is to 
 make a first order Taylor series expansion of~\eqref{eqn:proofloggaussian} with respect to the new variables 
\begin{subequations}
\label{eqn:transformvariables}
\begin{align}
Z\triangleq& X^{-1},\\
N^j\triangleq& (y^j-Hx)(y^j-Hx)^\t,\qquad j=1,\ldots,m,
\end{align}
\end{subequations}
around  the corresponding nominal values
\begin{subequations}
\label{eqn:transformnominal}
\begin{align}
\widehat{Z}\triangleq&\widehat{X}^{-1}\\
\widehat{N}^j\triangleq&(y^j-H\hat{x})(y^j-H\hat{x})^\t,\qquad j=1,\ldots,m.
\end{align}
\end{subequations}
Writing~\eqref{eqn:proofloggaussian} in terms of $Z$ and $N=[N^1, \dots, N^m]$, we obtain
\begin{align}
-&2\sum_{j=1}^m\log\N(y^j;Hx,sX+R)\pluseq m\log |Z^{-1}| \nonumber\\
&+m\log \left|sI+Z^{1/2}RZ^{1/2}\right|+\sum_{j=1}^m \tr\left[N^j(sZ^{-1}+R)^{-1}\right]\label{eqn:proofloggaussian1}
\end{align}
If we make a first order Taylor series expansion for the second and the third terms on the right hand side of~\eqref{eqn:proofloggaussian1}
with respect to the variables $Z$ and $N$ around $\widehat{Z}$ and $\widehat{N}$, we obtain
\begin{align}
-&2\sum_{j=1}^m\log\N(y^j;Hx,sX+R)\plusapprox \nonumber\\
 &m\log |Z^{-1}|+m\tr\left((sR^{-1}+\widehat{Z})^{-1}Z\right)\nonumber\\
&+\sum_{j=1}^m \tr \left[{N^j(s\widehat{Z}^{-1}+R)^{-1}}\right]\nonumber\\
&+s\sum_{j=1}^m \tr\left[\widehat{Z}^{-1}(s\widehat{Z}^{-1}+R)^{-1}\widehat{N}^j(s\widehat{Z}^{-1}+R)^{-1}\widehat{Z}^{-1}Z\right] \label{eqn:proofloggaussian2}
\end{align}
where $\plusapprox$ represents an approximation up to an additive constant. The first order Taylor series approximations for the scalar valued functions of matrix variables of interest are given in Appendix~\ref{sec:appendixDerivations}. We now substitute back the relations~\eqref{eqn:transformvariables} and~\eqref{eqn:transformnominal}  into~\eqref{eqn:proofloggaussian2} to obtain the approximation
\begin{align}
-2\sum_{j=1}^m\log&\N(y^j;Hx,sX+R)\nonumber\\
\plusapprox&m\log |X|+m\tr\left((sR^{-1}+\widehat{X}^{-1})^{-1}X^{-1}\right)\nonumber\\
&+\sum_{j=1}^m\tr\left[(y^j-Hx)(y^j-Hx)^\t (s\widehat{X}+R)^{-1}\right] \nonumber\\
&+s\sum_{j=1}^m\tr\left[\widehat{X}(s\widehat{X}+R)^{-1}(y^j-H\hat{x})(y^j-H\hat{x})^\t\right.\nonumber\\
&\left.\times(s\widehat{X}+R)^{-1}\widehat{X}X^{-1}\right]\\
=&m\log |X|+m\tr\left((sR^{-1}+\widehat{X}^{-1})^{-1}X^{-1}\right)\nonumber\\
&+\sum_{j=1}^m(y^j-Hx)^\t (s\widehat{X}+R)^{-1} (y^j-Hx)\nonumber\\
&+s\tr\left[\widehat{X}(s\widehat{X}+R)^{-1}\sum_{j=1}^m(y^j-H\hat{x})(y^j-H\hat{x})^\t \right.\nonumber \\
&\left.\times(s\widehat{X}+R)^{-1}\widehat{X}X^{-1}\right]\label{eqn:proofloggaussian3}
\end{align}
Rearranging the terms in~\eqref{eqn:proofloggaussian3}, dividing both sides by $-2$ and then taking the exponential of both sides, we can see that
\begin{align}
\prod_{j=1}^m&\N(y^j;Hx,sX+R)\nonumber\\
&\timesapprox \left(\prod_{j=1}^m \N(y^j;Hx,s\widehat{X}+R)\right)\IW\left(X;m,M\right)
\end{align}
where the sign $\timesapprox$ denotes an approximation up to a multiplicative constant and $M$ is given as
\begin{align}
&M\triangleq m\left(sR^{-1}+\widehat{X}^{-1}\right)^{-1}+s\widehat{X}(s\widehat{X}+R)^{-1}\nonumber\\
&\times\left(\sum_{j=1}^m(y^j-H\hat{x})(y^j-H\hat{x})^\t\right) (s\widehat{X}+R)^{-1}\widehat{X}.
\end{align}
Another form for the inverse Wishart parameter $M$ can be written using the matrix inversion lemma as follows.
\begin{align}
M=&m\widehat{X}+ms\widehat{X}(s\widehat{X}+R)^{-1}\nonumber\\
&\times\left[\left(\frac{1}{m}\sum_{j=1}^m(y^j-H\hat{x})(y^j-H\hat{x})^\t\right)-(s\widehat{X}+R)\right] \nonumber\\
&\times(s\widehat{X}+R)^{-1}\widehat{X}
\end{align}
Proof is complete.
\subsection{First Order Taylor Series Approximations for Some Scalar Valued Functions of Matrix Variables}
\label{sec:appendixDerivations}
In this section, some first order Taylor series approximations of some scalar valued functions of
matrix arguments will be studied. The two functions we consider are given as
\begin{itemize}
\item Function 1
\begin{align}
f_1(Z)\triangleq\log \left|sI+Z^{1/2}RZ^{1/2}\right|
\end{align}
\item Function 2
\begin{align}
f_2(Z)\triangleq\tr\left(N(sZ^{-1}+R)^{-1}\right)
\end{align}
\end{itemize}
Both functions can be approximated with a first order Taylor series approximation around a nominal point $\widehat{Z}$ as follows.
\begin{align}
f_k(Z)\approx f_k(\widehat{Z})+\tr\left(F_k^T(Z-\widehat{Z})\right)
\end{align}
for $k=1,2$ where $F_k$ is defined as
\begin{align}
[F_k]_{ij}\triangleq\left.\frac{\partial f_k}{\partial z_{ij}}\right|_{Z=\widehat{Z}}
\end{align}
for $1\le i,j\le d$ where the notation $[\cdot]_{ij}$ denotes the element corresponding to $i$th row and $j$th column of the argument matrix and $z_{ij}\triangleq [Z]_{ij}$. Hence, in order to construct the required Taylor series approximation, we are only required to calculate the matrices $F_1$ and $F_2$.
\subsubsection{Calculation of $F_1$}
We first write $f_1(\cdot)$ as
\begin{align}
f_1(Z)\triangleq&\log \left|sI+Z^{1/2}RZ^{1/2}\right|\\
=&\log |Z|+ \log \left|sZ^{-1}+R\right|
\end{align}
Now, we can calculate the related derivative using  two well-known matrix derivatives 
\begin{align}
\frac{\partial \ln|\mathbf{U}|}{\partial x} =&	{\rm tr}\left(\mathbf{U}^{-1}\frac{\partial \mathbf{U}}{\partial x}\right),\\
\frac{\partial \mathbf{U}^{-1}}{\partial x} =& 	-\mathbf{U}^{-1} \frac{\partial \mathbf{U}}{\partial x}\mathbf{U}^{-1}.
\end{align}
\begin{align}
\frac{\partial f_1}{\partial z_{ij}}=&\tr\left(Z^{-1}\frac{\partial Z}{\partial z_{ij}}\right)\nonumber\\
&+\tr\left((sZ^{-1}+R)^{-1}\frac{\partial(sZ^{-1}+R)}{\partial z_{ij}} \right)\\
=&\tr\left(Z^{-1}E_{ij}\right)+s\tr\left((sZ^{-1}+R)^{-1}\frac{\partial Z^{-1}}{\partial z_{ij}} \right)\\
=&[Z^{-1}]_{ji}-s\tr\left((sZ^{-1}+R)^{-1}Z^{-1}\frac{\partial Z}{\partial z_{ij}}Z^{-1} \right)\\
=&[Z^{-1}]_{ji}-s\tr\left(Z^{-1}(sZ^{-1}+R)^{-1}Z^{-1}E_{ij}\right)\\
=&[Z^{-1}]_{ji}-s\left[Z^{-1}(sZ^{-1}+R)^{-1}Z^{-1}\right]_{ji}\\
=&\left[Z^{-1}-sZ^{-1}(sZ^{-1}+R)^{-1}Z^{-1}\right]_{ji}\\
=&\left[(Z+sR^{-1})^{-1}\right]_{ji}\label{eqn:derivationF1}
\end{align}
where $E_{ij}$ is a $d\times d$ matrix filled with zeros except the $ij$th element which is unity. The last expression~\eqref{eqn:derivationF1} is equivalent to
\begin{align}
F_1=(Z+sR^{-1})^{-\t{}}
\end{align}
which gives
\begin{align}
\log \left|sI+Z^{1/2}RZ^{1/2}\right|  \approx \log& \left|sI+R\widehat{Z}\right| \nonumber\\
&+\tr\left((\widehat{Z}+sR^{-1})^{-1}(Z-\widehat{Z})\right)
\end{align}

\subsubsection{Calculation of $F_2$}
\begin{align}
&\frac{\partial f_2}{\partial z_{ij}}=\tr\left(N\frac{\partial (sZ^{-1}+R)^{-1}}{\partial z_{ij}} \right)\\
=&\tr\left(N\frac{\partial (sZ^{-1}+R)^{-1}}{\partial z_{ij}} \right)\\
=&-\tr\Bigl(N(sZ^{-1}+R)^{-1}\frac{\partial (sZ^{-1}+R)}{\partial z_{ij}}\nonumber\\
&\times(sZ^{-1}+R)^{-1} \Bigr)\\
=&-s\tr\left(N(sZ^{-1}+R)^{-1}\frac{\partial Z^{-1}}{\partial z_{ij}}(sZ^{-1}+R)^{-1} \right)\\
=&s\tr\Bigl(N(sZ^{-1}+R)^{-1}Z^{-1}\frac{\partial Z}{\partial z_{ij}}Z^{-1}\nonumber\\
&\times(sZ^{-1}+R)^{-1} \Bigr)\\
=&s\tr\left(N(sZ^{-1}+R)^{-1}Z^{-1}E_{ij}Z^{-1}(sZ^{-1}+R)^{-1}\right)\\
=&s\tr\left(Z^{-1}(sZ^{-1}+R)^{-1}N(sZ^{-1}+R)^{-1}Z^{-1}E_{ij}\right)\\
=&s\left[Z^{-1}(sZ^{-1}+R)^{-1}N(sZ^{-1}+R)^{-1}Z^{-1}\right]_{ji}\label{eqn:derivationF2}
\end{align}
 The last expression~\eqref{eqn:derivationF1} is equivalent to
\begin{align}
F_2=s\left[Z^{-1}(sZ^{-1}+R)^{-1}N(sZ^{-1}+R)^{-1}Z^{-1}\right]^{\t{}}
\end{align}
which gives
\begin{align}
&\tr\left(N(sZ^{-1}+R)^{-1}\right)\approx\tr\left(N(s\widehat{Z}^{-1}+R)^{-1}\right)\nonumber\\
&+s\tr\left(\widehat{Z}^{-1}(s\widehat{Z}^{-1}+R)^{-1}N(s\widehat{Z}^{-1}+R)^{-1}\widehat{Z}^{-1}(Z-\widehat{Z})\right)
\end{align}

\end{document}